\documentclass[runningheads]{llncs}

\usepackage{eccv}

\usepackage{eccvabbrv}

\usepackage{graphicx}
\usepackage{booktabs}
\usepackage{multirow}

\usepackage[accsupp]{axessibility}  

\usepackage{xcolor}
\usepackage{pgffor}
\usepackage{array}
\usepackage{tabularx}
\usepackage{wrapfig,lipsum}
\newcolumntype{Y}{>{\centering\arraybackslash}X}
\newcolumntype{x}[1]{>{\centering\arraybackslash\hspace{0pt}}p{#1}}

\usepackage{hyperref}

\usepackage{orcidlink}
\usepackage{adjustbox}

\newcommand{\coolname}{\textit{EgoLifter}}

\newcommand{\supp}{Appendix}

\begin{document}

\title{\coolname: Open-world 3D Segmentation for Egocentric Perception} 


\author{Qiao Gu\inst{1,2}\thanks{Work done during internship at Reality Labs, Meta.}\orcidlink{0000-0002-1342-3307} \and 
Zhaoyang Lv\inst{2}\orcidlink{0000-0002-7788-9982} \and 
Duncan Frost\inst{2}\orcidlink{0000-0002-5948-3452} \and 
Simon Green\inst{2}\orcidlink{0009-0006-5323-1170} \and 
\\Julian Straub\inst{2}\orcidlink{0000-0003-2339-1262} \and 
Chris Sweeney\inst{2}\orcidlink{0009-0005-5318-5621}}

\authorrunning{Q.~Gu et al.}

\institute{University of Toronto, Toronto, ON M5S 1A1, Canada \\ \email{q.gu@mail.utoronto.ca} \and Meta Reality Labs, Redmond, WA 98052, USA\\
\email{\{zhaoyang, frost, simongreen, jstraub, sweeneychris\}@meta.com}}

\maketitle

\begin{abstract}
In this paper we present~\coolname, a novel system that can automatically segment scenes captured from egocentric sensors into a complete decomposition of individual 3D objects. The system is specifically designed for egocentric data where scenes contain hundreds of objects captured from natural (non-scanning) motion. \coolname~adopts 3D Gaussians as the underlying representation of 3D scenes and objects and uses segmentation masks from the Segment Anything Model (SAM) as weak supervision to learn flexible and promptable definitions of object instances free of any specific object taxonomy. To handle the challenge of dynamic objects in ego-centric videos, we design a transient prediction module that learns to filter out dynamic objects in the 3D reconstruction. The result is a fully automatic pipeline that is able to reconstruct 3D object instances as collections of 3D Gaussians that collectively compose the entire scene. We created a new benchmark on the Aria Digital Twin dataset that quantitatively demonstrates its state-of-the-art performance in open-world 3D segmentation from natural egocentric input. We run \coolname~on various egocentric activity datasets which shows the promise of the method for 3D egocentric perception at scale. Please visit project page at \href{https://egolifter.github.io/}{https://egolifter.github.io/}. 
    \vspace{-0.5em}
  \vspace{-0.5em}
\end{abstract}

\section{Introduction}
\label{sec:intro}

\begin{figure}[th!]
  \centering
  \includegraphics[width=\linewidth,trim={0.2in 8.6in 5.5in 0},clip]{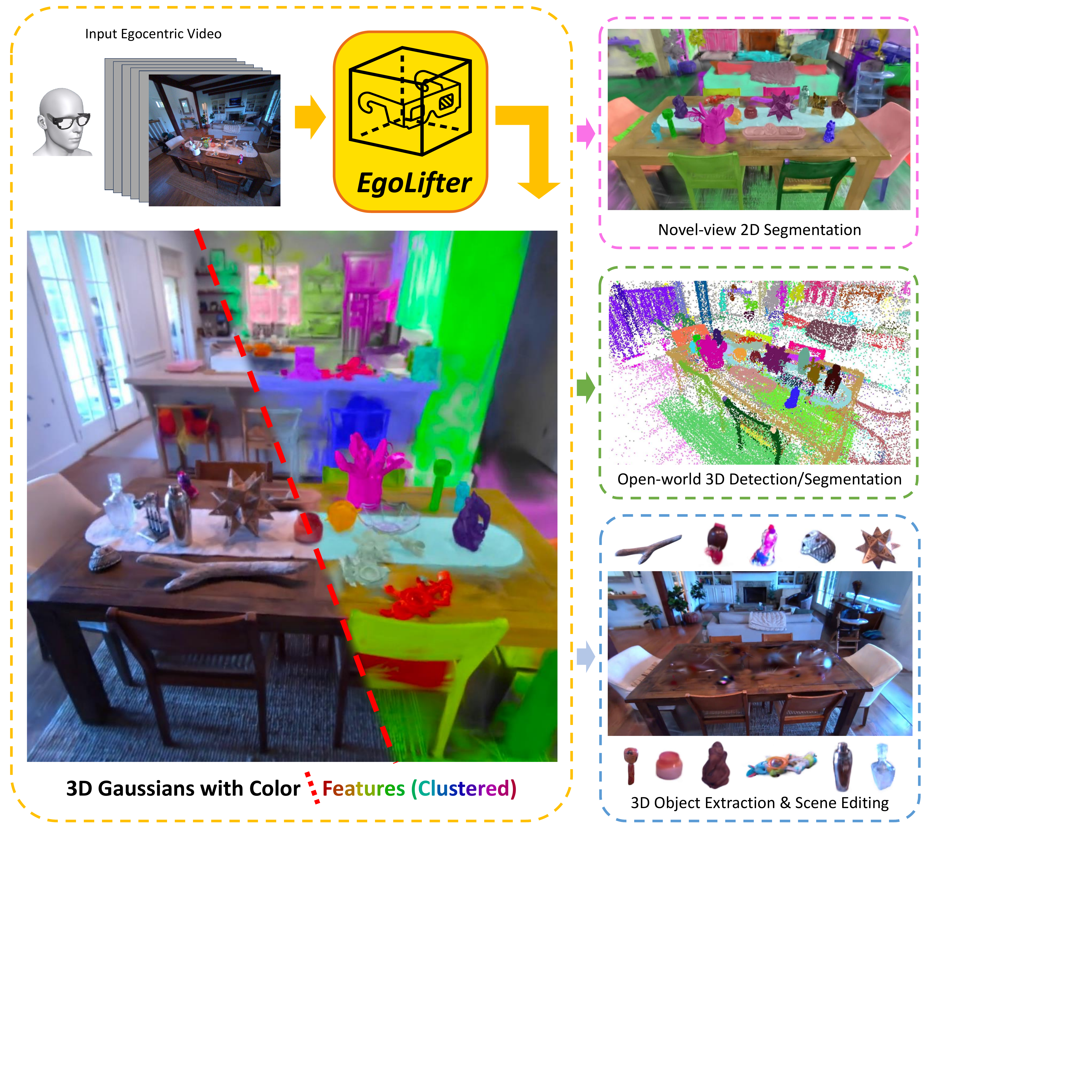}
  \caption{\coolname~solves 3D reconstruction and open-world segmentation simultaneously from egocentric videos. \coolname~augments 3D Gaussian Splatting~\cite{kerbl2023gaussiansplatting} with instance features and lifts open-world 2D segmentation by contrastive learning, where 3D Gaussians belonging to the same objects are learned to have similar features. In this way, \coolname~solves the multi-view mask association problem and establishes a consistent 3D representation that can be decomposed into object instances. \coolname~enables multiple downstream applications including detection, segmentation, 3D object extraction and scene editing. See \href{https://egolifter.github.io/}{project webpage} for animated visualizations.
  }
  \vspace{-1.5em}
  \label{fig:teaser}
  \vspace{-1em}
\end{figure}

The rise of personal wearable devices has led to the increased importance of egocentric machine perception algorithms capable of understanding the physical 3D world around the user.  Egocentric videos directly reflect the way humans see the world and contain important information about the physical surroundings and how the human user interacts with them. The specific characteristics of egocentric motion, however, present challenges for 3D computer vision and machine perception algorithms. Unlike datasets captured with deliberate "scanning" motions, egocentric videos are not guaranteed to provide complete coverage of the scene. This makes reconstruction challenging due to limited or missing multi-view observations. 
The specific content found in egocentric videos also presents challenges to conventional reconstruction and perception algorithms. An average adult interacts with hundreds of different objects many thousands of times per day~\cite{crabtree2016day}.  
Egocentric videos capturing this frequent human-object interaction thus contain a huge amount of dynamic motion with challenging occlusions. A system capable of providing useful scene understanding from egocentric data must therefore be able to recognize hundreds of different objects while being robust to sparse and rapid dynamics.

To tackle the above challenges, we propose~\coolname, a novel egocentric 3D perception algorithm that simultaneously solves reconstruction and open-world 3D instance segmentation from egocentric videos. We represent the geometry of the scene using 3D Gaussians~\cite{kerbl2023gaussiansplatting} that are trained to minimize photometric reconstruction of the input images. To learn a flexible decomposition of objects that make up the scene we leverage SAM~\cite{kirillov2023sam} for its strong understanding of objects in 2D and lift these object priors into 3D using contrastive learning. Specifically, 3D Gaussians are augmented with additional N-channel feature embeddings that are rasterized into feature images. These features are then learned to encode the object segmentation information by contrastive lifting~\cite{bhalgat2023contrastivelift}. This technique allows us to learn a flexible embedding with useful object priors that can be used for several downstream tasks.

To handle the difficulties brought by the dynamic objects in egocentric videos, we design \coolname~to focus on reconstructing the static part of the 3D scene. \coolname~learns a transient prediction network to filter out the dynamic objects from the reconstruction process. This network does not need extra supervision and is optimized together with 3D Gaussian Splatting using solely the photometric reconstruction losses. We show that the transient prediction module not only helps with photorealistic 3D reconstruction but also results in cleaner lifted features and better segmentation performance.

\coolname~is able to reconstruct a 3D scene while decomposing it into 3D object instances without the need for any human annotation. The method is evaluated on several egocentric video datasets. The experiments demonstrate strong 3D reconstruction and open-world segmentation results. We also showcase several qualitative applications including 3D object extraction and scene editing. 
The contributions of this paper can be summarized as follows:
\begin{itemize}
    \item We demonstrate~\coolname, the first system that can enable open-world 3D understanding from natural dynamic egocentric videos.
    \item By lifting output from recent image foundation models to 3D Gaussian Splatting, \coolname~achieve strong open-world 3D instance segmentation performance without the need for expensive data annotation or extra training. 
    \item We propose a transient prediction network, which filters out transient objects from the 3D reconstruction results. By doing so, we achieve improved performance on both reconstruction and segmentation of static objects. 
    \item We set up the first benchmark of dynamic egocentric video data and quantitatively demonstrate the leading performance of \coolname. On several large-scale egocentric video datasets, \coolname~showcases the ability to decompose a 3D scene into a set of 3D object instances, which opens up promising directions for egocentric video understanding in AR/VR applications. 
\end{itemize}

\section{Related Work}
\label{sec:related}

\subsection{3D Gaussian Models}

3D Gaussian Splatting (3DGS)~\cite{kerbl2023gaussiansplatting} has emerged as a powerful algorithm for novel view synthesis by 3D volumetric neural rendering. It has shown promising performance in many applications, like 3D content generation~\cite{tang2023dreamgaussian, chen2023text, yi2023gaussiandreamer}, SLAM~\cite{yan2023gsslam, matsuki2023gaussian, keetha2023splatam} and autonomous driving~\cite{yan2024street, zhou2023drivinggaussian}. 
Recent work extend 3DGS to dynamic scene reconstruction~\cite{luiten2023dynamic, yang2023deformable, wu20234d, duan20244d, yang2023real}. 
The pioneering work from Luiten~\etal~\cite{luiten2023dynamic} first learns a static 3DGS using the multi-view observations at the initial timestep and then updates it by the observations at the following timesteps. 
Later work~\cite{wu20234d, yang2023deformable} reconstructs dynamic scenes by deforming a canonical 3DGS using a time-conditioned deformation network. 
Another line of work~\cite{duan20244d, yang2023real} extends 3D Gaussians to 4D, with an additional variance dimension in time. While they show promising results in dynamic 3D reconstruction, they typically require training videos from multiple static cameras. 
However, in egocentric perception, there are only one or few cameras with a narrow baseline. As we show in the experiments, dynamic 3DGS struggles to track dynamic objects and results in floaters that harm instance segmentation feature learning. 

\subsection{Open-world 3D Segmentation}

Recent research on open-world 3D segmentation~\cite{liu20233dovs, conceptfusion, peng2023openscene, tschernezki2022n3f, kobayashi2022ffd, clipfields, tsagkas2023vlfields, lerf2023, huang23avlmaps, shen2023distilled, Engelmann2023openreno, mazur2023feature, gu2022ossid} has focused on lifting outputs from 2D open-world models - large, powerful models that are trained on Internet-scale datasets and can generalize to a wide range of concepts~\cite{clip, gpt4, kirillov2023sam, stablediffusion}. 
These approaches transfer the ability of powerful 2D models to 3D, require no training on 3D models, and alleviate the need for large-scale 3D datasets that are expensive to collect. 
Early work~\cite{peng2023openscene, conceptfusion, lerf2023} lifts dense 2D feature maps to 3D representations by multi-view feature fusion, where each position in 3D is associated with a feature vector. This allows queries in fine granularity over 3D space, but it also incurs high memory usage. 
Other work~\cite{gu2023conceptgraphs, takmaz2023openmask3d, lu2023ovir3d} builds object-decomposed 3D maps using 2D open-world detection or segmentation models~\cite{liu2023grounding, kirillov2023sam}, where each 3D object is reconstructed separately and has a single feature vector. This approach provides structured 3D scene understanding in the form of object maps or scene graphs but the scene decomposition is predefined and the granularity does not vary according to the query at inference time. 
Recently, another work~\cite{bhalgat2023contrastivelift} lifts 2D instance segmentation to 3D by contrastive learning. It augments NeRF~\cite{mildenhall2021nerf} with an extra feature map output and optimizes it such that pixels belonging to the same 2D segmentation mask are pulled closer and otherwise pushed apart. In this way, multi-view association of 2D segmentation is solved in an implicit manner and the resulting feature map allows instance segmentation by either user queries or clustering algorithms. 

\vspace{-1.5em}
\subsubsection{Concurrent Work.}
We briefly review several recent and unpublished pre-prints that further explore topics in this direction using techniques similar to ours. Concurrently, OmniSeg3D~\cite{ying2023omniseg3d} and GARField~\cite{kim2024garfield} follow the idea of~\cite{bhalgat2023contrastivelift}, and focus on learning 3D hierarchical segmentation. They both take advantage of the multi-scale outputs from SAM~\cite{kirillov2023sam} and incorporate the scales into the lifted features. GaussianGrouping~\cite{ye2023gaussiangrouping} also approaches the open-world 3D segmentation problem but they rely on a 2D video object tracker for multi-view association instead of directly using 2D segmentation via contrastive learning. Similar to our improvement on 3DGS, FMGS~\cite{zuo2024fmgs}, LangSplat~\cite{qin2023langsplat} and Feature3DGS~\cite{zhou2024feature3dgs} also augment 3DGS with feature rendering. They learn to embed the dense features from foundation models~\cite{clip, oquab2023dinov2} into 3DGS such that the 3D scenes can be segmented by language queries. While the concurrent work collectively also achieves 3D reconstruction with the open-world segmentation ability, \coolname~is the first to explicitly handle the dynamic objects that are commonly present in the real-world and especially in egocentric videos. We demonstrate this is a challenge in real-world scenarios and show improvements on it brought by \coolname.

\subsection{3D Reconstruction from Egocentric Videos}

NeuralDiff~\cite{tschernezki2021neuraldiff} first approached the problem of training an egocentric radiance field reconstruction by decomposing NeRF into three branches, which capture ego actor, dynamic objects, and static background respectively as inductive biases. 
EPIC-Fields~\cite{tschernezki2023epicfields} propose an augmented benchmark using 3D reconstruction by augmenting the EPIC-Kitchen~\cite{damen2018epickitchen} dataset using neural reconstruction. They also provide comprehensive reconstruction evaluations of several baseline methods~\cite{tschernezki2021neuraldiff, martin2021nerfw, gao2022monocular}. 
Recently, two datasets for egocentric perception, Aria Digital Twin (ADT) dataset~\cite{pan2023adt} and Aria Everyday Activities (AEA) Dataset~\cite{lv2024aea}, have been released. Collected by Project Aria devices~\cite{engel2023aria}, both datasets feature egocentric video sequences with human actions and contain multimodal data streams and high-quality 3D information. ADT also provides extensive ground truth annotations using a motion capture system. 
Preliminary studies on egocentric 3D reconstruction have been conducted on these new datasets~\cite{sun2023aria, lv2024aea} and demonstrate the challenges posed by dynamic motion. 
In contrast, this paper tackles the challenges in egocentric 3D reconstruction and proposes to filter out transient objects in the videos. 
Compared to all existing work, we are the first work that holistically tackles the challenges in reconstruction and open-world scene understanding, and set up the quantitative benchmark to systematically evaluate performance in egocentric videos.

\section{Method}
\label{sec:method}

\newcommand{\bI}{\mathbf{I}}
\newcommand{\bM}{\mathbf{M}}
\newcommand{\bp}{\mathbf{p}}
\newcommand{\bP}{\mathbf{P}}
\newcommand{\bbf}{\mathbf{f}}
\newcommand{\gauss}{\mathbf{\Theta}}
\newcommand{\cam}{\theta}
\newcommand{\featmap}{\mathbf{F}}
\newcommand{\loss}{\mathcal{L}}
\newcommand{\maskset}{\mathcal{M}}
\newcommand{\pixelset}{\mathcal{U}}

\begin{figure}[tb]
  \centering
  \includegraphics[width=\linewidth,trim={0 2.8in 0.9in 0},clip]{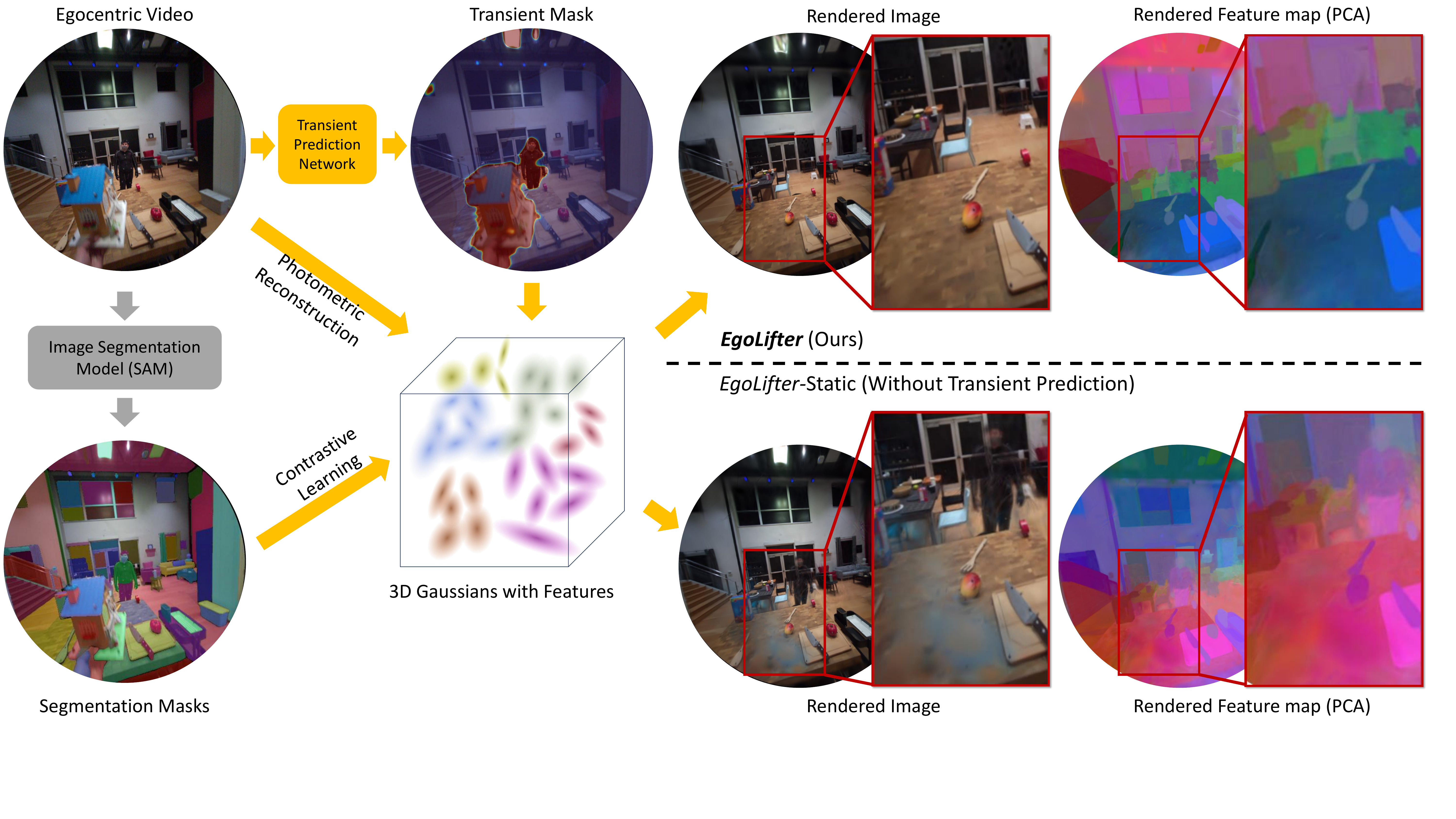}
  \caption{Naive 3D reconstruction from egocentric videos creates a lot of "floaters" in the reconstruction and leads to blurry rendered images and erroneous instance features (bottom right). 
  \coolname~tackles this problem using a transient prediction network, which predicts a probability mask of transient objects in the image and guides the reconstruction process. In this way, \coolname~gets a much cleaner reconstruction of the static background in both RGB and feature space (top right), which in turn leads to better object decomposition of 3D scenes. 
  }
  \label{fig:pipeline}
  \vspace{-1.5em}
\end{figure}

\subsection{3D Gaussian Splatting with Feature Rendering}

3D Gaussian Splatting (3DGS)~\cite{kerbl2023gaussiansplatting} has shown state-of-the-art results in 3D reconstruction and novel view synthesis. However, the original design only reconstructs the color radiance in RGB space and is not able to capture the rich semantic information in a 3D scene. In \coolname, we augment 3DGS to also render a feature map of arbitrary dimension in a differentiable manner, which enables us to encode high-dimensional features in the learned 3D scenes and lift segmentation from 2D to 3D. These additional feature channels are used to learn object instance semantics in addition to photometric reconstruction.

Formally, 3DGS represents a 3D scene by a set of $N$ colored 3D Gaussians $\mathcal{S}=\{\gauss_i | i=1,\cdots,N\}$, with location and shape represented by a center position $\bp_i \in \mathbb{R}^3$, an anisotropic 3D covariance $\mathbf{s}_i \in \mathbb{R}^3$ and a rotation quaternion $\mathbf{q}_i \in \mathbb{R}^4$. The radiance of each 3D Gaussian is described by an opacity parameter $\alpha_i \in \mathbb{R}$ and a color vector $\mathbf{c}_i$, parameterized by spherical harmonics (SH) coefficients. 
In \coolname, we additionally associate each 3D Gaussian with an extra feature vector $\bbf \in \mathbb{R}^d$, and thus the optimizable parameter set for $i$-th Gaussian is $\gauss_i=\{\bp_i, \mathbf{s}_i, \mathbf{q}_i, \alpha_i, \mathbf{c}_i, \bbf_i\}$.

To train 3DGS for 3D reconstruction, a set of $M$ observations $\{\bI_j, \cam_j | j=1,\cdots,M\}$ is used, where $\bI_j$ is an RGB image and $\cam_j$ is the corresponding camera parameters. 
During the differentiable rendering process, all 3D Gaussians are splatted onto the 2D image plane according to $\cam_j$ and $\alpha$-blended to get a rendered image $\hat\bI_j$. 
Then the photometric loss is computed between the rendered image $\hat\bI_j$ and the corresponding ground truth RGB image $\bI_j$ as
\begin{equation}\label{eq:loss-rgb}
    \loss_{\text{RGB}}(\bI_j, \hat\bI_j) = \loss_{\text{MSE}} (\bI_j, \hat\bI_j) = \sum_{u \in \Omega} \|\bI_j[u] - f(\hat\bI_j[u])\|_2^2,
\end{equation}
where $\loss_{\text{MSE}}$ is the mean-squared-error (MSE) loss, $\Omega$ is set of all coordinates on the image and $\bI_j[u]$ denotes the pixel value of $\bI_j$ at coordinate $u$.
$f(\cdot)$ is an image formation model that applies special properties of the camera (e.g. vignetting, radius of valid pixels) on the rendered image. 
By optimizing $\loss_{RGB}$, the location, shape, and color parameters of 3D Gaussians are updated to reconstruct the geometry and appearance of the 3D scene. A density control mechanism is also used to split or prune 3D Gaussians during the training process~\cite{kerbl2023gaussiansplatting}.

In \coolname, we also implement the differentiable feature rendering pipeline similar to that of the RGB images, which renders to a 2D feature map $\hat\featmap \in \mathbb{R}^{H\times W\times d}$ according to the camera information. During the training process, the feature vectors are supervised by segmentation output obtained from 2D images and jointly optimized with the location and color parameters of each Gaussian. We also include gradients of feature learning for the density control process in learning 3DGS. More details may be found in the supplementary material.  

\subsection{Learning Instance Features by Contrastive Loss}

Egocentric videos capture a huge number of different objects in everyday activities, and some of them may not exist in any 3D datasets for training. Therefore, egocentric 3D perception requires an ability to generalize to unseen categories (open-world) which we propose to achieve by lifting the output from 2D instance segmentation models. 
The key insight is that 2D instance masks from images of different views can be associated to form a consistent 3D object instance and that this can be done together with the 3D reconstruction process. Recent work has approached this problem using linear assignment~\cite{siddiqui2023panoptic}, video object tracking~\cite{ye2023gaussiangrouping}, and incremental matching~\cite{gu2023conceptgraphs, lu2023ovir3d, takmaz2023openmask3d}. 

To achieve open-world 3D segmentation, we use $\bbf$ as instance features to capture the lifted segmentation and their similarity to indicate whether a set of Gaussians should be considered as the same object instance. Inspired by Contrastive Lift~\cite{bhalgat2023contrastivelift}, we adopt supervised contrastive learning, which pulls the rendered features belonging to the same mask closer and pushes those of different masks further apart. 
Formally, given a training image $\bI_j$, we use a 2D segmentation model to extract a set of instance masks $\maskset_j=\{\bM_j^k | k=1,\cdots,m_i\}$ from $\bI_j$. 
The feature map $\hat\featmap_j$ at the corresponding camera pose $\cam_j$ is then rendered, and the contrastive loss is computed over a set of pixel coordinates $\pixelset$, for which we use a uniformly sampled set of pixels $\pixelset \subset \Omega$ due to GPU memory constraint. The contrastive loss is formulated as
\begin{equation}
    \loss_{\text{contr}} (\hat\featmap_j, \maskset_j) = - \frac{1}{|\pixelset|} \sum_{u \in \pixelset} \log 
    \frac{\sum_{u' \in \pixelset^+} \exp(\text{sim}(\hat\featmap_j[u], \hat\featmap_j[u']; \gamma)}
    {\sum_{u' \in \pixelset} \exp(\text{sim}(\hat\featmap_j[u], \hat\featmap_j[u']; \gamma)}, 
\end{equation}
where $\pixelset^+$ is the set of pixels that belong to the same instance mask as $u$ and $\hat\featmap_j[u]$ denotes the feature vector of the $\hat\featmap_j$ at coordinate $u$. We use a Gaussian RBF kernel as the similarity function, i.e. $\text{sim}(f_1, f_2; \gamma) = \text{exp}(-\gamma\|f_1 - f_2\|_2^2)$. 

In the contrastive loss, pixels on the same instance mask are considered as positive pairs and will have similar features during training. Note that since the 2D segmentation model does not output consistent object instance IDs across different views, the contrastive loss is computed individually on each image. This weak supervision allows the model to maintain a flexible definition of object instances without hard assignments and is key to learning multi-view consistent instance features for 3D Gaussians that enables flexible open-world 3D segmentation.

\subsection{Transient Prediction for Egocentric 3D Reconstruction}

Egocentric videos contain a lot of dynamic objects that cause many inconsistencies among 3D views. As we show in \cref{fig:pipeline}, the original 3DGS algorithm on the egocentric videos results in many floaters and harms the results of both reconstruction and feature learning. In \coolname, we propose to filter out transient phenomena in the egocentric 3D reconstruction, by predicting a transient probability mask from the input image, which is used to guide the 3DGS reconstruction process. 

Specifically, we employ a transient prediction network $G(\bI_j)$, which takes in the training image $\bI_j$ and outputs a probability mask $\bP_j \in \mathbb{R}^{H\times W}$ whose value indicates the probability of each pixel being on a transient object. Then $\bP_j$ is used to weigh the reconstruction loss during training, such that when a pixel is considered transient, it is filtered out in reconstruction. Therefore the reconstruction loss from \cref{eq:loss-rgb} is adapted to
\begin{equation}\label{eq:loss-rgb-w}
    \loss_{\text{RGB-w}} (\bI_j, \hat\bI_j, \bP_j) = \sum_{u \in \Omega} (1 - \bP_j[u]) \|\bI_j[u] - \hat\bI_j[u]\|_2^2,
\end{equation}
where the pixels with lower transient probability will contribute more to the reconstruction loss. 
As most of the objects in egocentric videos remain static, we also apply an $L$-1 regularization loss on the predicted $\bP_j$ as
$\loss_{\text{reg}}(\bP_j) = \sum_{p\in \bP_j} | p |.$
This regularization also helps avoid the trivial solution where $\bP_j$ equals one and all pixels are considered transient. The transient mask $\bP_j$ is also used to guide contrastive learning for lifting instance segmentation, where the pixel set $\pixelset$ is only sampled on pixels with the probability of being transient less than a threshold $\delta$. As shown in \cref{fig:pipeline} and \cref{fig:qual-image-feat}, this transient filtering also helps learn cleaner instance features and thus better segmentation results. 

In summary, the overall training loss on image $\bI_j$ is a weighted sum as
\begin{equation}\label{eq:loss-all}
    \loss = \lambda_1 \loss_{\text{RGB-w}} (\bI_j, \hat\bI_j, \bP_j) +
    \lambda_2 \loss_{\text{contr}} (\hat\featmap_j, \maskset_j)  + 
    \lambda_3 \loss_{\text{reg}}(\bP_j),
\end{equation}
with $\lambda_1$, $\lambda_2$ and $\lambda_3$ as hyperparameters. 

\subsection{Open-world Segmentation}
\label{sec:open-world-seg}

After training, instance features $\bbf$ capture the similarities among 3D Gaussians, and can be used for open-world segmentation in two ways, query-based and clustering-based. In query-based open-world segmentation, one or few clicks on the object of interest are provided and a query feature vector is computed as the averaged features rendered at these pixels. Then a set of 2D pixels or a set of 3D Gaussians can be obtained by thresholding their Euclidean distances from the query feature, from which a 2D segmentation mask or a 3D bounding box can be estimated. In clustering-based segmentation, an HDBSCAN clustering algorithm~\cite{mcinnes2017hdbscan} is performed to assign 3D Guassians into different groups, which gives a full decomposition of the 3D scene into a set of individual objects. In our experiments, query-based segmentation is used for quantitative evaluation, and clustering-based mainly for qualitative results.

\section{Experiments}
\label{sec:experiment}

\subsubsection*{Implementation.} We use a U-Net~\cite{ronneberger2015unet} with the pretrained MobileNet-v3~\cite{howard2019mobilenetv3} backbone as the transient prediction network $G$. The input to $G$ is first resized to $224\times224$ and then we resize its output back to the original resolution using bilinear interpolation. We use feature dimension $d=16$, threshold $\delta=0.5$, temperature $\gamma=0.01$, and loss weights $\lambda_1=1$, $\lambda_2=0.1$ and $\lambda_3=0.01$. The 3DGS is trained using the Adam optimizer~\cite{kingma2014adam} with the same setting and the same density control schedule as in~\cite{kerbl2023gaussiansplatting}. The transient prediction network is optimized by another Adam optimizer with an initial learning rate of $1\times10^{-5}$. 
\coolname~is agnostic to the specific 2D instance segmentation method, and we use the Segment Anything Model (SAM)~\cite{kirillov2023sam} for its remarkable instance segmentation performance.  

\vspace{-1.5em}
\subsubsection*{Datasets.} We evaluate \coolname~on the following egocentric datasets:
\begin{itemize}
    \item \textbf{Aria Digital Twin (ADT)~\cite{pan2023adt}} provides 3D ground truth for objects paired with egocentric videos, which we used to evaluate \coolname~quantitatively. ADT dataset contains 200 egocentric video sequences of daily activities, captured using Aria glasses. ADT also uses a high-quality simulator and motion capture devices for extensive ground truth annotations, including 3D object bounding boxes and 2D segmentation masks for all frames. 
    ADT does not contain an off-the-shelf setting for scene reconstruction or open-world 3D segmentation. We create the evaluation benchmark using the GT 2D masks and 3D bounding boxes by reprocessing the 3D annotations. Note that only the RGB images are used during training, and for contrastive learning, we used the masks obtained by SAM~\cite{kirillov2023sam}. 
    \item \textbf{Aria Everyday Activities (AEA) dataset~\cite{lv2024aea}} provides 143 egocentric videos of various daily activities performed by multiple wearers in five different indoor locations. Different from ADT, AEA contains more natural video activity recordings but does not offer 3D annotations. For each location, multiple sequences of different activities are captured at different times but aligned in the same 3D coordinate space. Different frames or recordings may observe the same local space at different time with various dynamic actions, which represent significant challenges in reconstruction.
    We group all daily videos in each location and run \coolname~for each spatial environment. The longest aggregated video in one location (Location 2) contains 2.3 hours of video recording and a total of 170K RGB frames. The dataset demonstrates our method can not only tackle diverse dynamic activities, but also produce scene understanding at large scale in space and time.
    \item \textbf{Ego-Exo4D~\cite{grauman2023egoexo4d} dataset} is a large and diverse dataset containing over one thousand hours of videos captured simultaneously by egocentric and exocentric cameras. Ego-Exo4D videos capture humans performing a wide range of activities. 
    We qualitatively evaluate \coolname~on the egocentric videos of Ego-Exo4D.
\end{itemize}

We use the same process for all Project Aria videos. Since Aria glasses use fisheye cameras, we undistort the captured images first before training. We use the image formation function $f(\cdot)$ in \cref{eq:loss-rgb} to capture the vignetting effect and the radius of valid pixels, according to the specifications of the camera on Aria glasses. We use high-frequency 6DoF trajectories to acquire RGB camera poses and the semi-dense point clouds provided by each dataset through the Project Aria Machine Perception Services (MPS).

\vspace{-1.5em}
\subsubsection*{Baselines.} We compare \coolname~to the following baselines.

\begin{itemize}
    \item \textbf{SAM ~\cite{kirillov2023sam}} masks serve as input to \coolname. The comparison on segmentation between \coolname~and SAM shows the benefits of multi-view fusion of 2D masks. As we will discuss in \cref{sec:exp-qual-eval}, SAM only allows prompts from the same image (in-view query), while \coolname~enables segmentation prompts from different views (cross-view query) and 3D segmentation. 
    \item \textbf{Gaussian Grouping~\cite{ye2023gaussiangrouping}} also lifts the 2D segmentation masks into 3D Gaussians. Instead of the contrastive loss, Gaussian Grouping uses a video object tracker to associate the masks from different views and employs a linear layer for identity classification. Gaussian Grouping does not handle the dynamic objects in 3D scenes. 
\end{itemize}

\vspace{-2em}
\subsubsection*{Ablations.} We further provide two variants of \coolname~in particular to study the impact of reconstruction backbone.
\begin{itemize}
    \item \textbf{\coolname-Static} disabled the transient prediction network. A vanilla static 3DGS~\cite{kerbl2023gaussiansplatting} is learned to reconstruct the scene. We use the same method to lift and segment 3D features.
    \item \textbf{\coolname-Deform} uses a dynamic variant of 3DGS~\cite{yang2023deformable} instead of the transient prediction network to handle the dynamics in the scene. Similar to~\cite{yang2023deformable}, \coolname-Deform learns a canonical 3DGS and a network to predict the location and shape of each canonical 3D Gaussian at different timestamps. 
\end{itemize}

We also compare \coolname~with NeRF-based methods~\cite{tancik2023nerfstudio, ying2023omniseg3d} on the ADT dataset and evaluate \coolname~on non-egocentric datasets~\cite{replica, barron2022mip360}. Please see~\cref{sec:extra-results} for the results. 

\subsection{Benchmark Setup on ADT}\label{sec:exp-qual-eval}

\begin{table}[tb]
    \caption{Quantitative evaluation of 2D instance segmentation (measured in mIoU) and novel view synthesis (measured in PNSR) on the ADT dataset. The evaluations are conducted on the frames in the novel subset of each scene.}
    \label{tab:adt-2dseg}
    \centering
    \vspace{-0.5em}
    \begin{adjustbox}{max width=\textwidth}
    \begin{tabular}{l | p{0.09\textwidth}p{0.11\textwidth}p{0.09\textwidth} | p{0.09\textwidth}p{0.11\textwidth}p{0.09\textwidth} | p{0.09\textwidth}p{0.11\textwidth}p{0.09\textwidth}}
    \toprule
    Evaluation  & \multicolumn{3}{c|}{mIoU (In-view)} & \multicolumn{3}{c|}{mIoU (Cross-view)} & \multicolumn{3}{c}{PSNR} \\ 
    Object set    &   Static & Dynamic & All       &   Static & Dynamic & All  &  Static & Dynamic & All  \\ \midrule
    SAM~\cite{kirillov2023sam} & 54.51 & 32.77 & 50.69 & \multicolumn{1}{c}{-} &  \multicolumn{1}{c}{-} & \multicolumn{1}{c|}{-} & \multicolumn{1}{c}{-} & \multicolumn{1}{c}{-} & \multicolumn{1}{c}{-} \\
    Gaussian Grouping ~\cite{ye2023gaussiangrouping}  &   35.68 & 30.76 & 34.81 & 23.79 & 11.33 & 21.58 & 21.29 & 14.99 & 19.97 \\
    \coolname-Static   & 55.67 & \textbf{39.61} & 52.86 & 51.29 & 18.67 & 45.49 & 21.37 & 15.32 & 20.16 \\
    \coolname-Deform  & 54.23 & 38.62 & 51.49 & 51.10 & 18.02 & 45.22 & 21.16 & \textbf{15.39} & 19.93 \\
    \textbf{\coolname~(Ours)}   &   \textbf{58.15} & 37.74 & \textbf{54.57} & \textbf{55.27} & \textbf{19.14} & \textbf{48.84} & \textbf{22.14} & 14.37 & \textbf{20.28} \\
    \bottomrule
    \end{tabular}
    \end{adjustbox}
    \vspace{-1em}
\end{table}

We use the ADT dataset~\cite{pan2023adt} for the quantitative evaluation.
We use 16 video sequences from ADT, and the frames of each sequence are split into \textbf{seen} and \textbf{novel} subsets. The seen subset are used for training and validation, while the novel subset contains a chunk of consecutive frames separate from the seen subset and is only used for testing. The evaluation on the novel subset reflects the performance on novel views. 
The objects in each video sequence are also tagged \textbf{dynamic} and \textbf{static} according to whether they move. Each model is trained and evaluated on one video sequence separately. 
We evaluate the performance of the query-based open-world 2D instance segmentation and 3D instance detection tasks, as described in \cref{sec:open-world-seg}. 
For the Gaussian Grouping baseline~\cite{ye2023gaussiangrouping}, we use their learned identity encoding for extracting query features and computing similarity maps. Please refer to~\supp~for more details of the evaluation settings, the exact sequence IDs and splits we used. 

\vspace{-1.5em}
\subsubsection{Open-world 2D instance segmentation.}
We adopt two settings in terms of query sampling for 2D evaluation, namely \textbf{in-view} and \textbf{cross-view}. In both settings, a similarity map is computed between the query feature and the rendered feature image. The instance segmentation mask is then obtained by cutting off the similarity map using a threshold that maximizes the IoU with respect to the GT mask, which resembles the process of a human user adjusting the threshold to get the desired object. 
In the \textbf{in-view} setting, the query feature is sampled from one pixel on the rendered feature map in the same camera view. 
For a fair comparison, SAM~\cite{kirillov2023sam} in this setup takes in the rendered images from the trained 3DGS and the same query pixel as the segmentation prompt. For each prompt, SAM predicts multiple instance masks at different scales, from which we also use the GT mask to pick one that maximizes the IoU for evaluation. 
The \textbf{cross-view} setting follows the prompt propagation evaluation used in the literature~\cite{ren2022nvos, mirzaei2023spinnerf, ying2023omniseg3d, cen2024sa3dnerf}. We randomly sample $5$ pixels from the training images (in the seen subset) on the object, and their average feature is used as the query for segmentation on the novel subset. 
To summarize, the in-view setting evaluates how well features group into objects after being rendered into a feature map, and the cross-view setting evaluates how well the learned feature represents each object instance in 3D and how they generalize to novel views. 

\vspace{-1.5em}
\subsubsection{Open-world 3D instance detection.}
For 3D evaluation, we use the same query feature obtained in the above cross-view setting. The similarity map is computed between the query feature and the learned 3D Gaussians, from which a subset of 3D Gaussians are obtained by thresholding, and a 3D bounding box is estimated based on their coordinates. The 3D detection performance is evaluated by the IoU between the estimated and the GT 3D bounding boxes. We also select the threshold that maximizes the IoU with the GT bounding box. We only evaluate the 3D static objects in each scene.

\vspace{-1.5em}
\subsubsection{Novel view synthesis.}
We evaluate the synthesized frames in the novel subset using PSNR metric. We use "All" to indicate full-frame view synthesis. We also separately evaluate the pixels on dynamic and static regions using the provided the 2D ground truth dynamic motion mask.

\begin{wraptable}{r}{3.9cm}
  \caption{3D instance detection performance for the static objects in the ADT dataset.}
  \label{tab:adt-3ddet}
  \centering
    \begin{adjustbox}{max width=0.3\textwidth}
    \begin{tabular}{p{0.3\textwidth}|l}
    \toprule
     Method & mIoU \\ \midrule
     Gaussian Grouping~\cite{ye2023gaussiangrouping} & 7.48  \\
     \coolname-Static & 21.10 \\
     \coolname-Deform & 20.58 \\
     \textbf{\coolname~(Ours)} & \textbf{23.11} \\
     \bottomrule
    \end{tabular}
    \end{adjustbox}
    \vspace{-1em}
\end{wraptable} 

\subsection{Quantitative Results on ADT}
\label{sec:quant-res}

The quantitative results are reported in \cref{tab:adt-2dseg} and \ref{tab:adt-3ddet}. 
As shown in \cref{tab:adt-2dseg}, \coolname~consistently outperforms all other baselines and variants in the reconstruction and segmentation of static object instances in novel views. Since transient objects are deliberately filtered out during training, \coolname~has slightly worse performance on dynamic objects, However, the improvements on static objects outweigh the drops on transient ones in egocentric videos and \coolname~still achieves the best overall results in all settings. Similarly, this trend also holds in 3D, and \coolname~has the best 3D detection performance as shown in \cref{tab:adt-3ddet}.

\subsection{Qualitative Results on Diverse Egocentric Datasets}
\label{sec:qual-eval}

\begin{figure}[tb]
    \centering
    \includegraphics[width=0.19\linewidth]{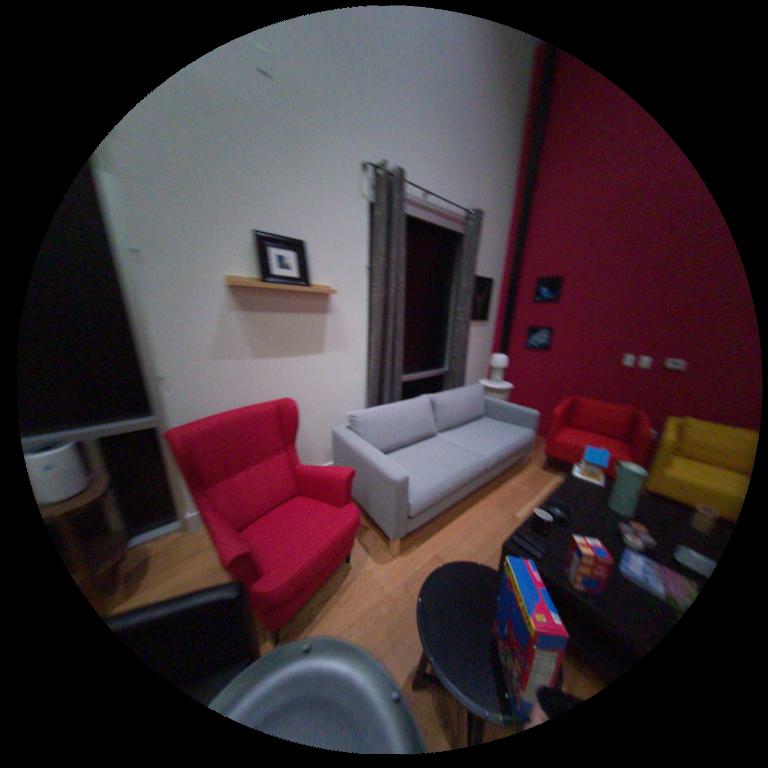}
    \includegraphics[width=0.19\linewidth]{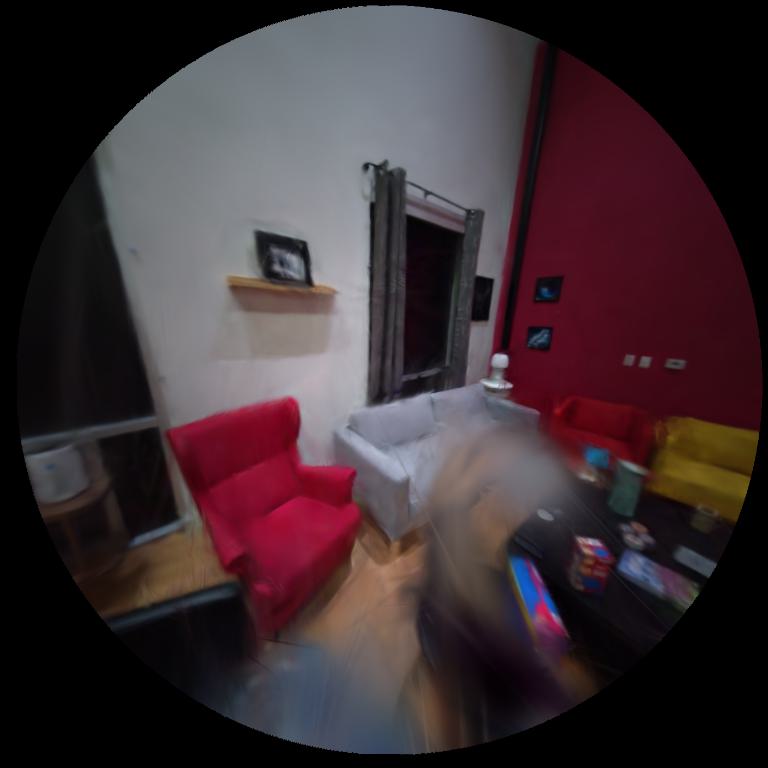}
    \includegraphics[width=0.19\linewidth]{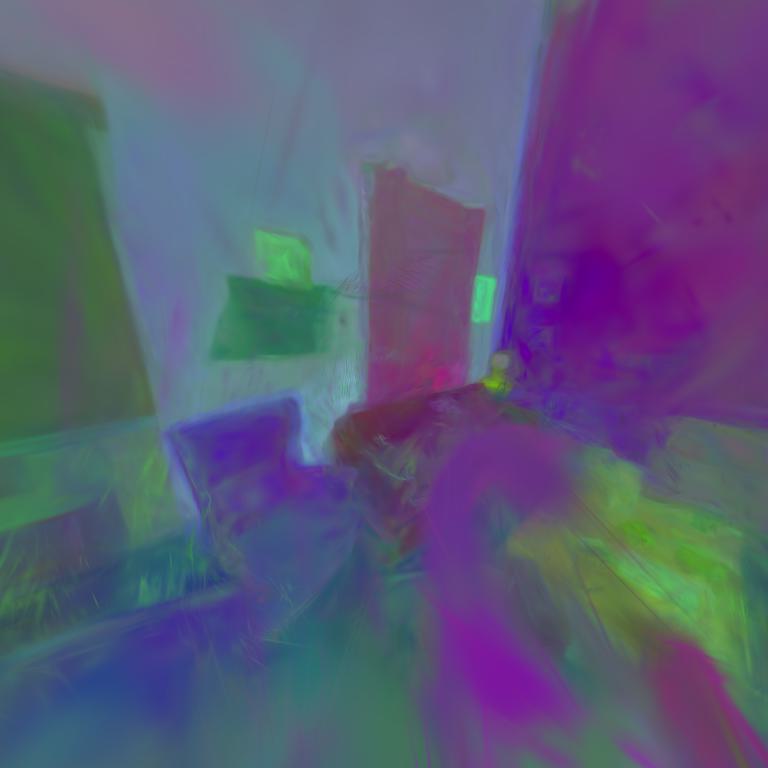}
    \includegraphics[width=0.19\linewidth]{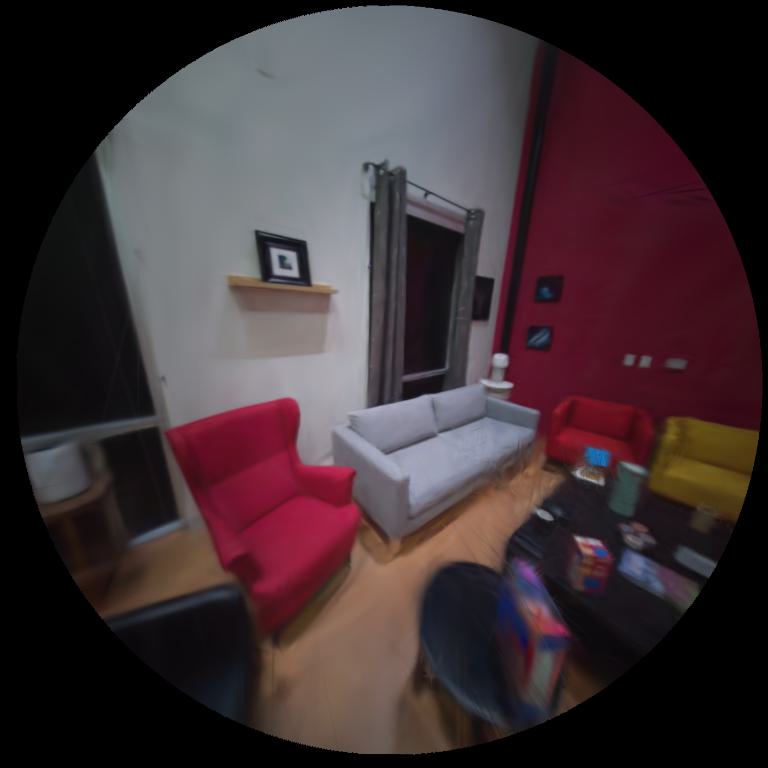}
    \includegraphics[width=0.19\linewidth]{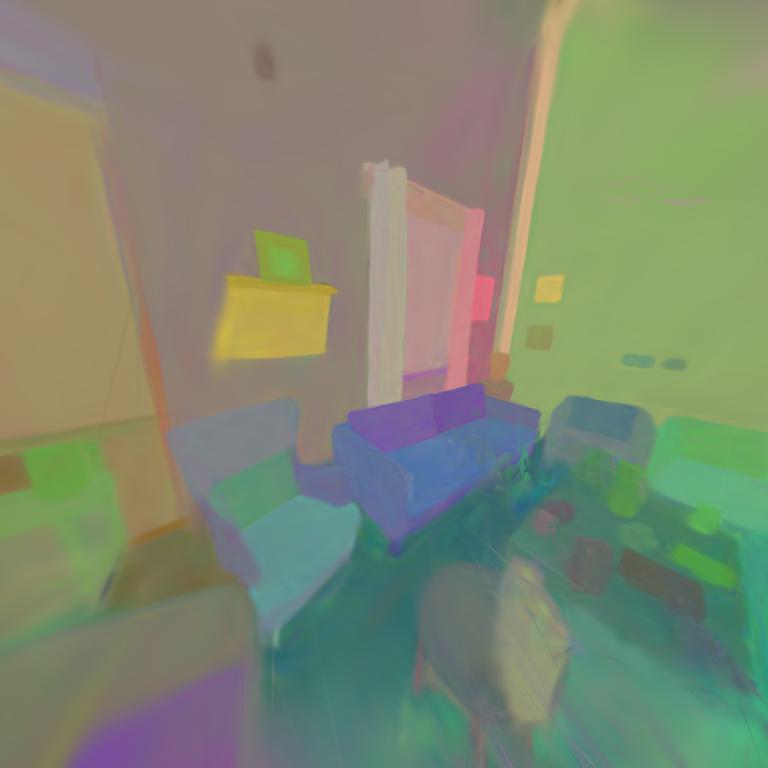}
    \begin{adjustbox}{max width=\textwidth}
    \begin{tabular}{x{0.25\textwidth} x{0.25\textwidth} x{0.25\textwidth} x{0.25\textwidth} x{0.25\textwidth}}
       GT  & Render from~\cite{ye2023gaussiangrouping} & Feature from~\cite{ye2023gaussiangrouping} & Our Render & Our Feature \\
    \end{tabular}
    \end{adjustbox}
    \caption{Rendered images and feature maps (visualised in PCA colors) by Gaussian Grouping~\cite{ye2023gaussiangrouping} and \coolname~(Ours).}
    \vspace{-0.5em}
    \label{fig:gg-compare}
\end{figure}

In \cref{fig:qual-image-feat}, we visualize the qualitative results on several egocentric datasets~\cite{pan2023adt, lv2024aea, grauman2023egoexo4d}.
Please refer to \href{https://egolifter.github.io/}{project webpage} for the videos rendered by \coolname~with comparison to baselines. 
As shown in \cref{fig:qual-image-feat}, without transient prediction, \coolname-Static creates 3D Gaussians to overfit the dynamic observations in some training views. However, since dynamic objects are not geometrically consistent and may be at a different location in other views, these 3D Gaussians become floaters that explain dynamics in a ghostly way, harming both the rendering and segmentation quality. 
In contrast, \coolname~correctly identifies the dynamic objects in each image using transient prediction and filters them out in the reconstruction. The resulting cleaner reconstruction leads to better results in novel view synthesis and segmentation, as we have already seen quantitatively in \cref{sec:quant-res}. 
We also compare the qualitative results with Gaussian Grouping~\cite{ye2023gaussiangrouping} in~\cref{fig:gg-compare}, from which we can see that Gaussian Grouping not only struggles with floaters associated with transient objects but also has a less clean feature map even on the static region. We hypothesize this is because our contrastive loss helps learn more cohesive identity features than the classification loss used in~\cite{ye2023gaussiangrouping}. This also explains why \coolname-Static significantly outperforms Gaussian Grouping in segmentation metrics as shown in \cref{tab:adt-2dseg} and \ref{tab:adt-3ddet}. 

\begin{figure}[p!]
    \footnotesize
    \def \qualList {
    {adt/0},
    {adt/1},
    {adt/2},
    {aea/2},
    {aea/4},
    {egoexo/1},
    {egoexo/0}%
    }
    \centering

    \newcommand{\tableContent}{
    }
    \foreach \qualName [count=\x from 1] in \qualList {
    \edef\temp{\noexpand\gdef\noexpand\tableContent{
    \tableContent 
    \includegraphics[width=0.158\textwidth]{figures/qual/\qualName_gt.jpg} &
    \includegraphics[width=0.158\textwidth]{figures/qual/\qualName_baseline_rgb.jpg} &
    \includegraphics[width=0.158\textwidth]{figures/qual/\qualName_baseline_feat.jpg} &
    \includegraphics[width=0.158\textwidth]{figures/qual/\qualName_ours_transient.jpg} &
    \includegraphics[width=0.158\textwidth]{figures/qual/\qualName_ours_rgb.jpg} &
    \includegraphics[width=0.158\textwidth]{figures/qual/\qualName_ours_feat.jpg}
    \\}}
    \temp
    }

    \begin{tabularx}{\textwidth}{c|cc|ccc}
      & \multicolumn{2}{c|}{\coolname-Static} & \multicolumn{3}{c}{\textbf{\coolname~(Ours)}} \\ 
     GT & Render & Feature & Trans. map & Render & Feature \\ \midrule
    \tableContent
    \end{tabularx}
    \vspace{-1em}
    \caption{RGB images and feature maps (colored by PCA) rendered by the \coolname~Static baseline and \coolname. The predicted transient maps (Trans. map) from \coolname~are also visualized, with red color indicating a high probability of being transient. Note that the baseline puts ghostly floaters on the region of transient objects, but \coolname~filters them out and gives a cleaner reconstruction of both RGB images and feature maps. Rows 1-3 are from ADT, rows 4-5 from AEA, and rows 6-7 from Ego-Exo4D.}
    \vspace{-1em}
    \label{fig:qual-image-feat}
\end{figure}

\subsection{3D Object Extraction and Scene Editing}
\label{sec:scene-edit}

\begin{figure}[tb]
  \centering
  \includegraphics[width=0.8\linewidth,trim={0 13in 12in 0},clip]{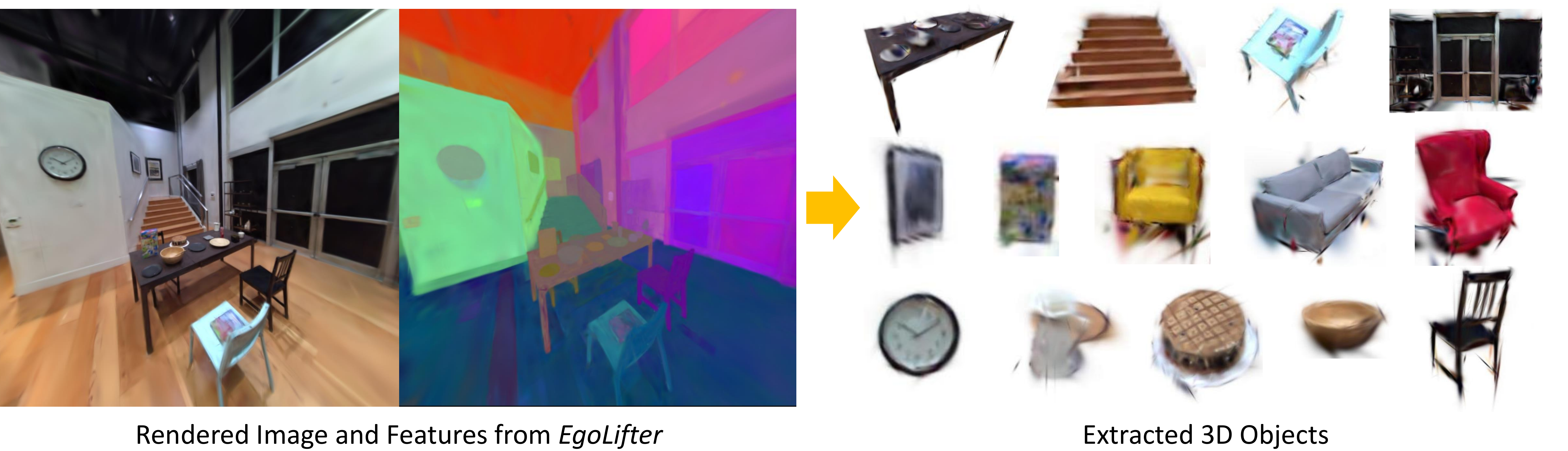}
  \caption{Individual 3D object can be extracted by querying or clustering over the 3D features from \coolname. Note object reconstructions are not perfect since each object might be partial observable in the egocentric videos rather than scanned intentionally. }
  \label{fig:scene-edit-vis}
  \vspace{-1.5em}
\end{figure}

Based on the features learned by \coolname, we can decompose a 3D scene into individual 3D objects, by querying or clustering over the feature space. Each extracted 3D object is represented by a set of 3D Gaussians which can be photo-realistically rendered. In \cref{fig:scene-edit-vis}, we show the visualization of 3D objects extracted from a scene in the ADT dataset. This can further enable scene editing applications by adding, removing, or transforming these objects over the 3D space.  In~\cref{fig:teaser}, we demonstrate one example of background recovery by removing all segmented 3D objects from the table. 

\section{Conclusion and Limitation}
\label{sec:conclusion}

We present \coolname, a novel algorithm that simultaneously solves the 3D reconstruction and open-world segmentation problem for in-the-wild egocentric perception. By lifting the 2D segmentation into 3D Gaussian Splatting, \coolname~achieves strong open-world 2D/3D segmentation performance with no 3D data annotation. To handle the rapid and sparse dynamics in egocentric videos, we employ a transient prediction network to filter out transient objects and get more accurate 3D reconstruction. \coolname~is evaluated on several challenging egocentric datasets and outperforms other existing baselines. The representations obtained by \coolname~can also be used for several downstream tasks like 3D object asset extraction and scene editing, showing great potential for personal wearable devices and AR/VR applications. 

\noindent\textbf{Limitations:}
We observe the transient prediction module may mix the regions that are hard to reconstruct with transient objects. As shown in rows (4) and (5) of \cref{fig:qual-image-feat}, the transient prediction module predicts a high probability for pixels on the windows, which have over-exposed pixels that are hard to be reconstructed from LDR images. In this case, \coolname~learns to filter them out to improve reconstruction on that region. Besides, the performance of \coolname~may also be dependent on the underlying 2D segmentation model. \coolname~is not able to segment an object if the 2D model consistently fails on it. 

\noindent\textbf{Potential Negative Impact:}
3D object digitization for egocentric videos in the wild may pose a risk to privacy considerations. 
Ownership of digital object rights of physical objects is also a challenging and complex topic that will have to be addressed as AR/VR becomes more ubiquitous.

\section*{Acknowledgments}

The authors thank the Project Aria team for providing open-source support and Nickolas Charron for helping with the ADT dataset evaluation. 
The authors also thank Fangzhou Hong, Kevin QH Lin, Hyo Jin Kim, Lingni Ma, Lambert Mathias, Pierre Moulon, Richard Newcombe, Tianwei Shen, Nan Yang, and Wang Zhao for discussions and useful feedback over the course of this project. 

\section*{Appendix}
\appendix

\setcounter{section}{0}
\setcounter{equation}{0}
\setcounter{figure}{0}
\setcounter{table}{0}

\renewcommand{\thesection}{A\arabic{section}}
\renewcommand{\thefigure}{A.\arabic{figure}}
\renewcommand{\thetable}{A.\arabic{table}}
\renewcommand{\theequation}{A.\arabic{equation}}

\newcommand{\fakeref}[1]{{\color{red}{#1}}}

\section{Video Qualitative Results}

Please refer to videos on the project page \footnote{\href{https://egolifter.github.io/}{https://egolifter.github.io/}}, which contains:
\begin{itemize}
    \item Videos qualitative results of the multiple applications of \coolname~(corresponding to Fig.~\fakeref{1}).
    \item Video qualitative results on the ADT dataset, comparing \coolname~and its variants (corresponding to Fig.~\fakeref{3}).
    \item Video qualitative results on the ADT dataset, comparing with Gaussian Grouping~\cite{ye2023gaussiangrouping} (corresponding to Fig.~\fakeref{4}).
    \item Video qualitative results on the AEA and Ego-Exo4D datasets. (corresponding to Fig.~\fakeref{3}).
    \item Demonstration video of the interactive visualization and segmentation system. 
\end{itemize}

\section{Experiment Details}

\subsection{Image Formation Model for Project Aria}

\begin{figure}
    \centering
    \includegraphics[width=\linewidth]{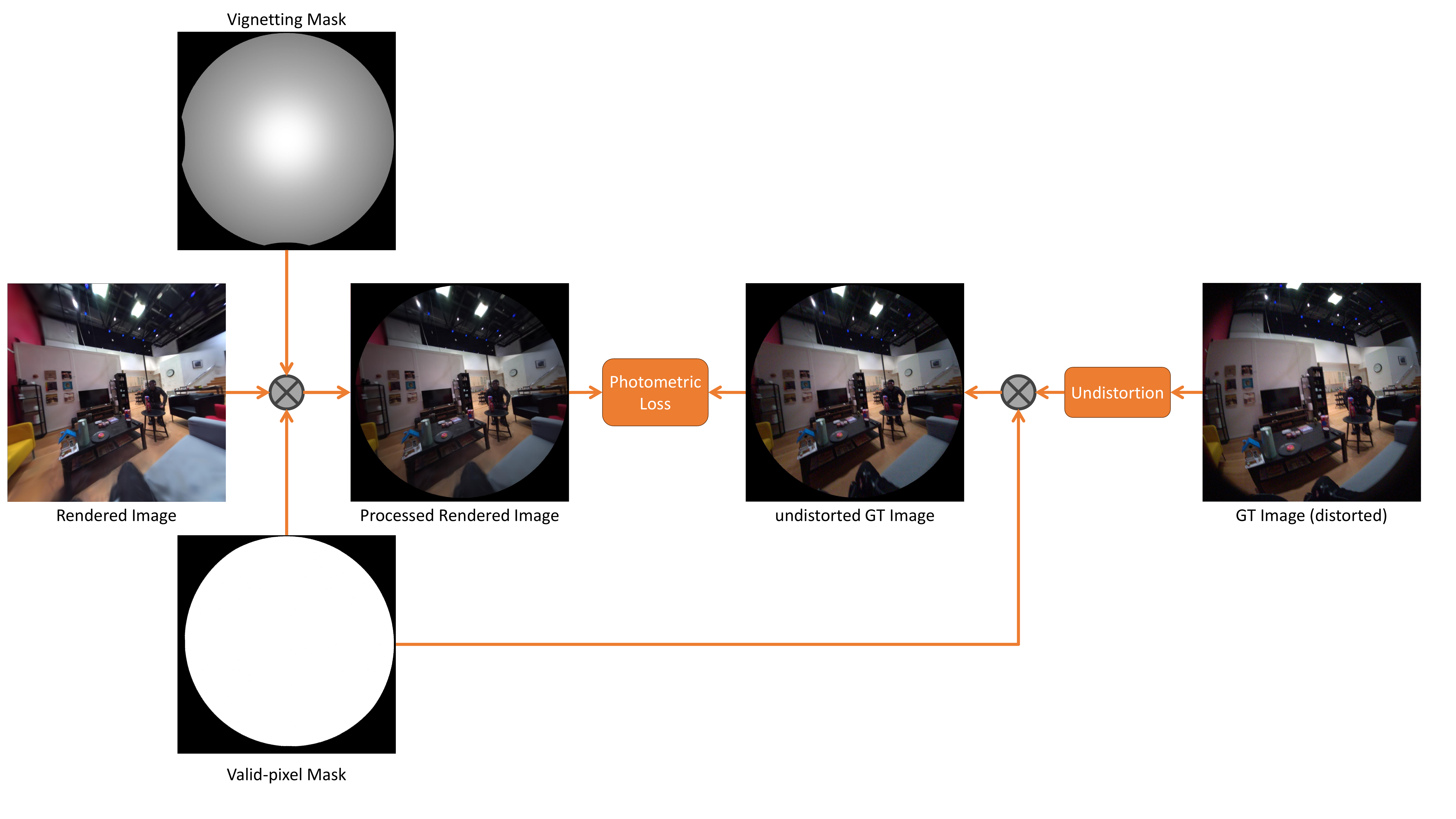}
    \caption{Image processing pipeline during training. The $\bigotimes$ symbol indicates element-wise multiplication. }
    \label{fig:image-process}
\end{figure}

Aria Glasses~\cite{engel2023aria} use a fisheye camera, and thus recorded images have a fisheye distortion and vignette effect, but 3DGS uses a linear camera model and does not have a vignette effect. Therefore we account for these effects in training 3D Gaussian models using the image formation model $f(\cdot)$ in Eq.~\fakeref{1}, such that not the raw rendered image but a processed one is used for loss computation. 
Specifically, we apply an image processing pipeline as shown in~\cref{fig:image-process}. In the pipeline, the raw recorded images are first rectified to a pinhole camera model using \verb|projectaria_tools|\footnote{\href{https://facebookresearch.github.io/projectaria\_tools/docs/data\_utilities/advanced\_code\_snippets/image\_utilities\#image-undistortion}{Link}}, and then multiplied with a valid-pixel mask that removes the pixels that are too far from the image center. The rendered image from 3DGS is multiplied with a vignette mask and also the valid-pixel masks. Then the photometric losses are computed between the processed rendered image and the processed GT image during training. This pipeline models the camera model used in Aria glasses and leads to better 3D reconstruction. Empirically we found that without this pipeline, 3DGS will create a lot of floaters to account for the vignette effect in the reconstruction and significantly harm the results. 


\subsection{Additional Training Details}

Due to the GPU memory constraint, we sampled at most $|\mathcal{U}|=4096$ pixels within the valid-pixel mask for computing the contrastive loss in Eq.~\fakeref{2}. Note that for \coolname~where the transient prediction is used, the samples are additionally constrained to be pixels with transient probability less than $\delta=0.5$. 

For the segmentation masks generated by SAM, some masks may have overlapped with each other. In our experiments, we discarded the information about overlapping and simply overlaid all masks on the image space to get a one-hot segmentation for each pixel. While making use of these overlapping results leads to interesting applications like hierarchical 3D segmentation as shown in~\cite{ying2023omniseg3d, kim2024garfield}, this is beyond the scope of \coolname~and we left this for future exploration. 
The images used for training are of resolution of $1408\times 1408$ and segmentation masks from SAM are in the resolution of $512\times 512$. Therefore, during training, two forward passes are performed. In the first pass, only the RGB image is rendered at the resolution of $1408\times 1408$ and in the second, only the feature map is rendered at $512\times 512$. The losses are computed separately from each pass and summed up for gradient computation. Note that the view-space gradients from both passes are also summed for deciding whether to split 3D Gaussians. 

For optimization on the 3D Gaussian models, we adopt the same setting as used in the original implementation~\cite{kerbl2023gaussiansplatting}, in terms of parameters used in the optimizer and scheduler and density control process. The learning rate for the additional per-Gaussian feature vector $\mathbf{f}_i$ is $0.0025$, the same as that for updating color $\mathbf{c}_i$. 
All models are trained for 30,000 iterations on each scene in the ADT dataset, and for 100,000 iterations on scenes in the AEA and Ego-Exo4D datasets, as these two datasets contain more frames in each scene. In the latter case, the learning rate scheduler and density control schedule are also proportionally extended. 

\subsection{Timing}

Using one NVIDIA A100 (40GB), training \coolname~on one ADT sequence takes around 130 minutes (training vanilla 3DGS takes around 100 minutes). For a trained \coolname~model, rendering both the RGB image and the instance feature map of $1408\times1408$ resolution runs at around 103 fps. If only RGB images are rendered, the speed goes to 158 fps. Note that we use a different implementation than the original 3DGS, where we made several changes like not caching images on GPU to enable training on large datasets, e.g. AEA and Ego-Exo4D. 

\begin{table}[tb]
    \caption{2D instance segmentation results (measured in mIoU) and novel view synthesis results (measured in PNSR) on \textbf{seen} subsets in the ADT dataset.}
    \label{tab:adt-2dseg-seen}
    \centering
    \vspace{-0.5em}
    \begin{adjustbox}{max width=\textwidth}
    \begin{tabular}{l | p{0.09\textwidth}p{0.11\textwidth}p{0.09\textwidth} | p{0.09\textwidth}p{0.11\textwidth}p{0.09\textwidth} | p{0.09\textwidth}p{0.11\textwidth}p{0.09\textwidth}}
    \toprule
    Evaluation  & \multicolumn{3}{c|}{mIoU (In-view)} & \multicolumn{3}{c|}{mIoU (Cross-view)} & \multicolumn{3}{c|}{PSNR} \\ 
    Object set    &   Static & Dynamic & All       &   Static & Dynamic & All  &  Static & Dynamic & All  \\ \midrule
    SAM~\cite{kirillov2023sam} & 62.74 & 52.48 & 61.00 & \multicolumn{1}{c}{-} &  \multicolumn{1}{c}{-} & \multicolumn{1}{c|}{-} & \multicolumn{1}{c}{-} & \multicolumn{1}{c}{-} & \multicolumn{1}{c}{-} \\
    Gaussian Grouping ~\cite{ye2023gaussiangrouping}  & 40.86 & 42.24 & 41.09 & 32.26 & 26.23 & 31.24 & 27.97 & 19.13 & 25.53 \\
    \coolname-Static   & 64.34 & \textbf{57.71} & \textbf{63.21} & 62.20 & \textbf{35.39} & 57.64 & 27.65 & 19.60 & 25.64 \\
    \coolname-Deform  & 63.33 & 57.11 & 62.27 & 62.24 & 34.91 & 57.59 & \textbf{28.60} & \textbf{19.89} & \textbf{26.24} \\
    \textbf{\coolname~(Ours)}   &   \textbf{65.08} & 52.12 & 62.88 & \textbf{63.65} & 33.70 & \textbf{58.56} & 26.86 & 16.02 & 23.34 \\
    \bottomrule
    \end{tabular}
    \end{adjustbox}
    \vspace{-1em}
\end{table}

\subsection{ADT Dataset Benchmark}

\subsubsection{Sequence selection}

Based on the 218 sequences in the full ADT datasets~\cite{pan2023adt}, we filter out the sequences that have too narrow baselines for 3D reconstruction (sequences with name starting with \verb|Lite_release_recognition|) or do not have segmentation annotation on human bodies. 
From the rest of the sequences, we select 16 sequences for evaluation, where 6 of them contain recordings of Aria glasses from two human users in the scene (sequences with \verb|multiskeleton| in the name), and the rest 10 only have recordings from one user, although there may be multiple two persons in the scene (sequences with \verb|multiuser| in the name). The names of the selected sequences are listed as follows:

\begin{verbatim}
Apartment_release_multiskeleton_party_seq121
Apartment_release_multiskeleton_party_seq122
Apartment_release_multiskeleton_party_seq123
Apartment_release_multiskeleton_party_seq125
Apartment_release_multiskeleton_party_seq126
Apartment_release_multiskeleton_party_seq127
Apartment_release_multiuser_cook_seq114
Apartment_release_multiuser_meal_seq140
Apartment_release_multiuser_cook_seq143
Apartment_release_multiuser_party_seq140
Apartment_release_multiuser_clean_seq116
Apartment_release_multiuser_meal_seq132
Apartment_release_work_skeleton_seq131
Apartment_release_work_skeleton_seq140
Apartment_release_meal_skeleton_seq136
Apartment_release_decoration_skeleton_seq137
\end{verbatim}

\subsection{Comparison with NeRF}

\begin{table}[b]
    \centering
    \begin{adjustbox}{max width=0.6\linewidth}
    \begin{tabular}{c|cccc}
    \toprule
    Method & Nerfacto & \coolname \\
    \midrule
    PSNR (all)   & 17.22 & \textbf{20.28}  \\
    \bottomrule
    \end{tabular}
    \end{adjustbox}
    \caption{Comparison to Nerfacto on the ADT dataset.}
    \label{tab:adt-nerf}
\end{table}

\begin{table}[t]
    \centering
    \begin{adjustbox}{max width=0.7\linewidth}
    \begin{tabular}{c|cccc}
    \toprule
    Method & INGP-Big &  M-NeRF360 & 3DGS & \coolname \\
    \midrule
    PSNR   & 25.59 & 27.69 & 27.21 & 27.26  \\
    \bottomrule
    \end{tabular}
    \end{adjustbox}
    \caption{Quantitative comparison on the MipNeRF 360 dataset.}
    \label{tab:mip360-quant}
\end{table}

\begin{figure}[th!]
    \centering
    \includegraphics[width=0.28\linewidth]{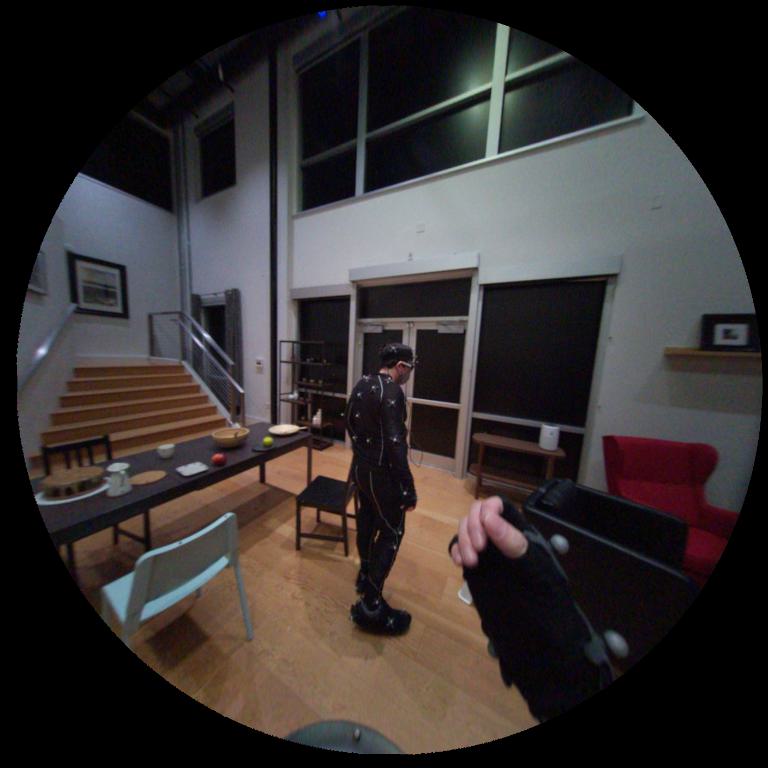}
    \includegraphics[width=0.28\linewidth]{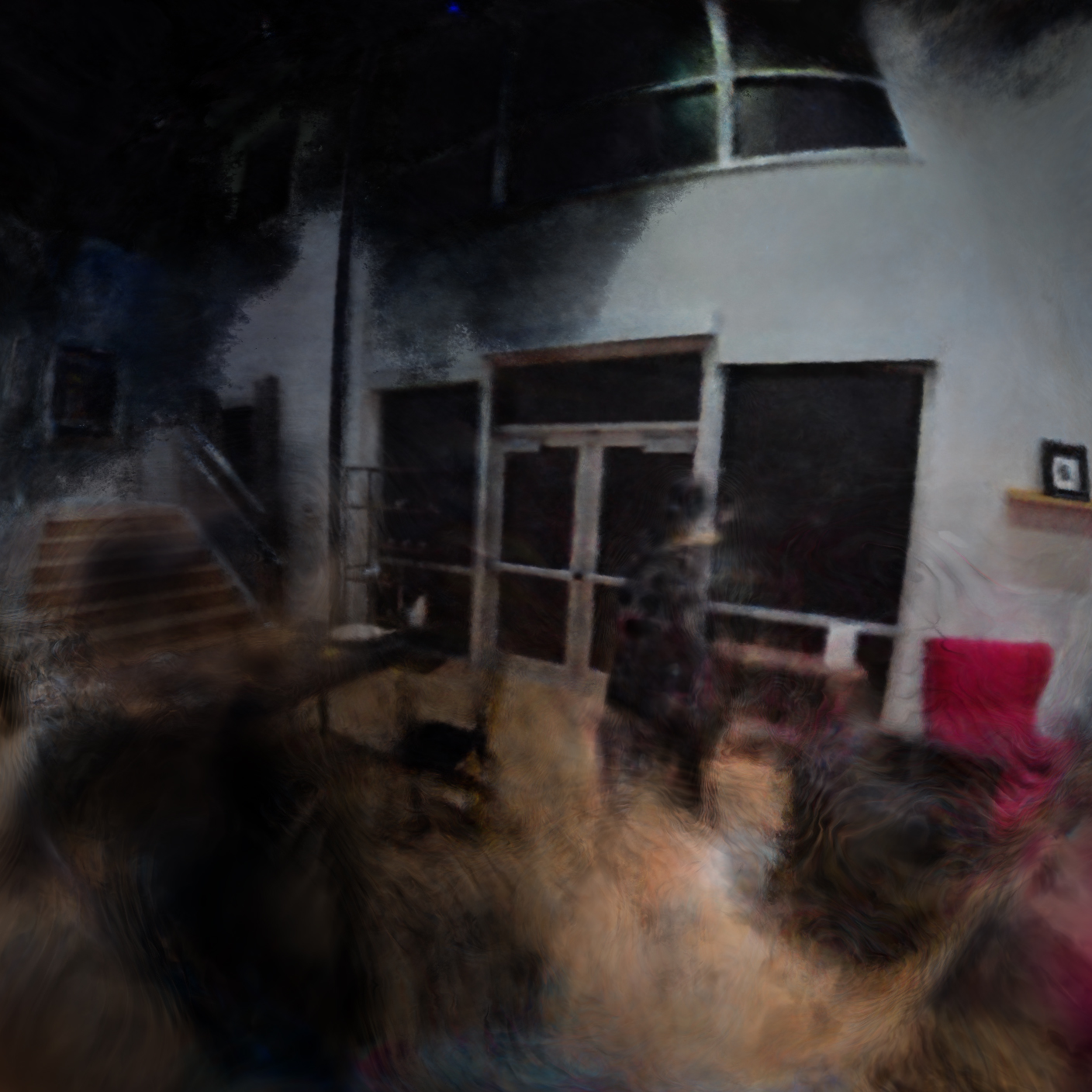}
    \includegraphics[width=0.28\linewidth]{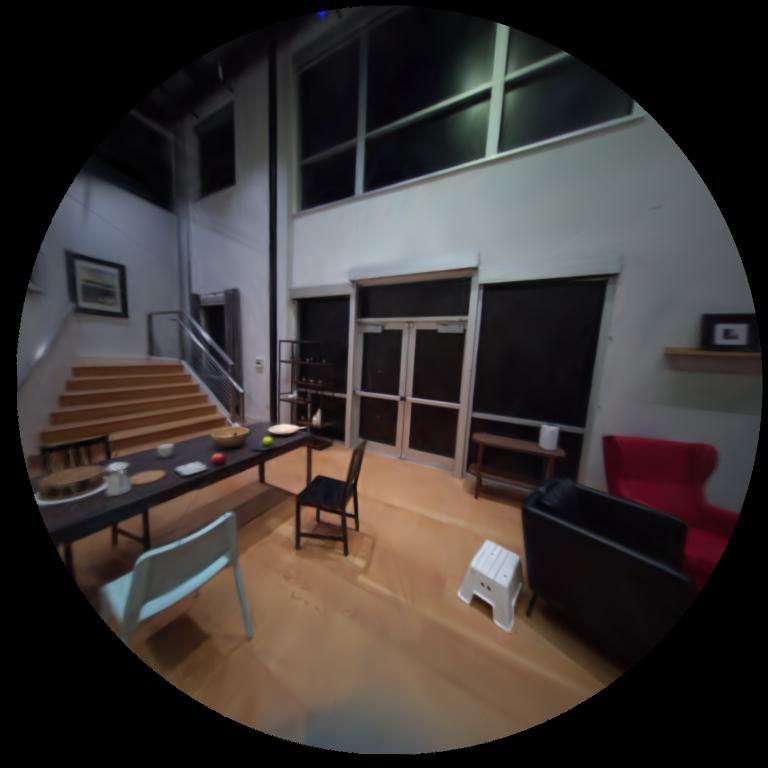}
    \includegraphics[width=0.28\linewidth]{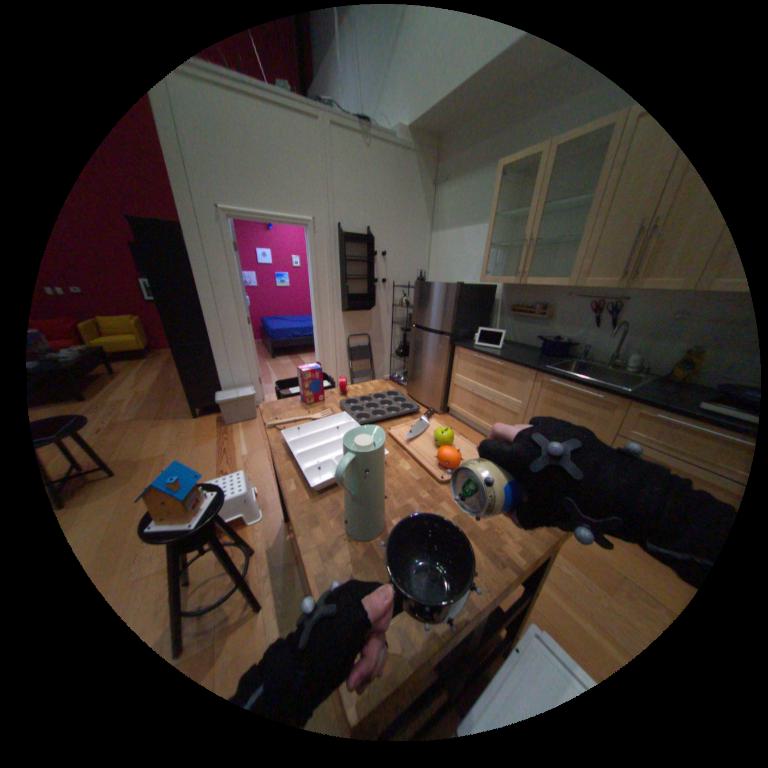}
    \includegraphics[width=0.28\linewidth]{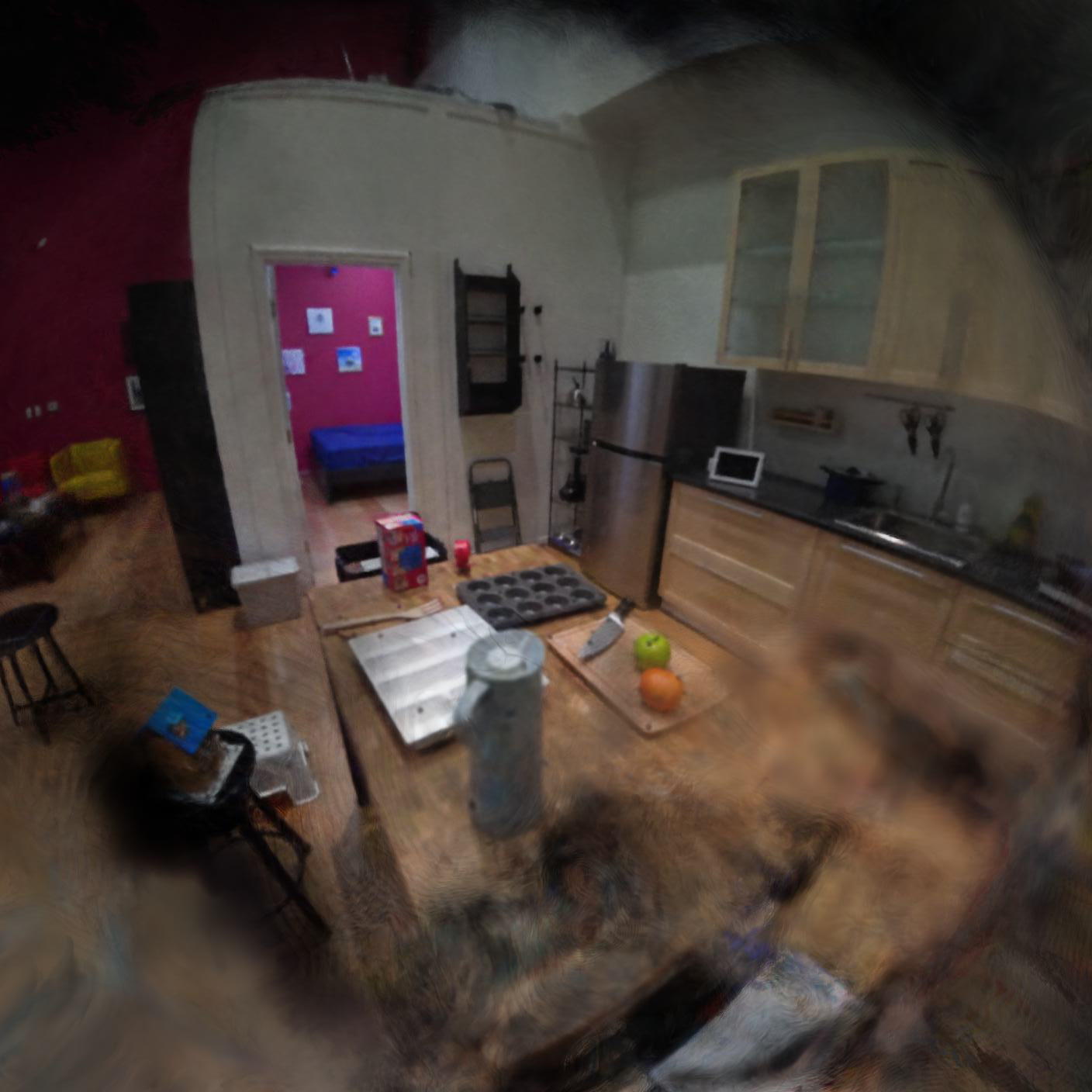}
    \includegraphics[width=0.28\linewidth]{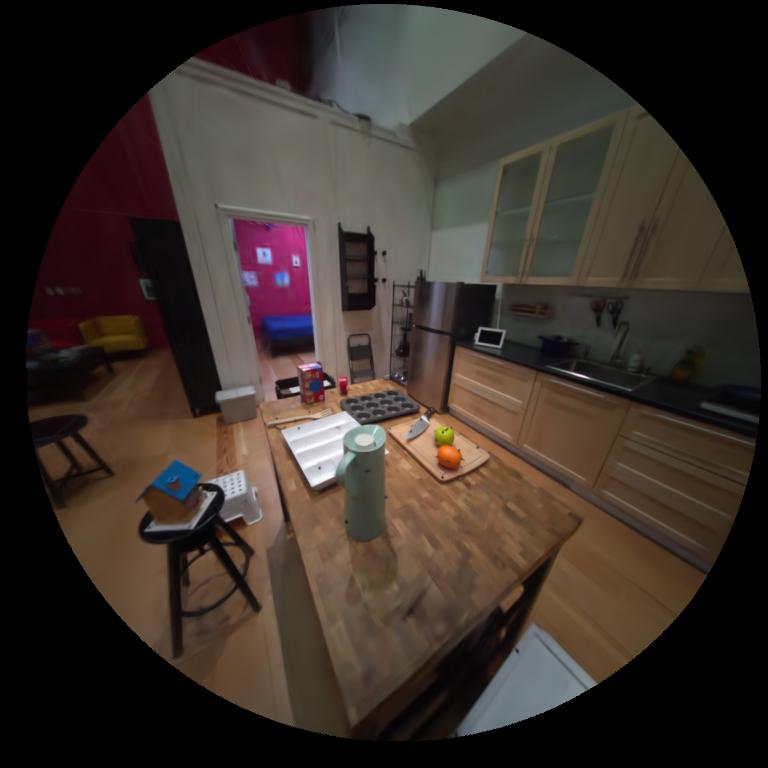}
    \caption{Qualitative results on ADT datasets. From left to right: GT image; Render by Nerfacto; Render by \coolname.}
    \label{fig:adt-qual}
\end{figure}

\begin{figure}[t]
    \centering
    \includegraphics[width=0.325\linewidth]{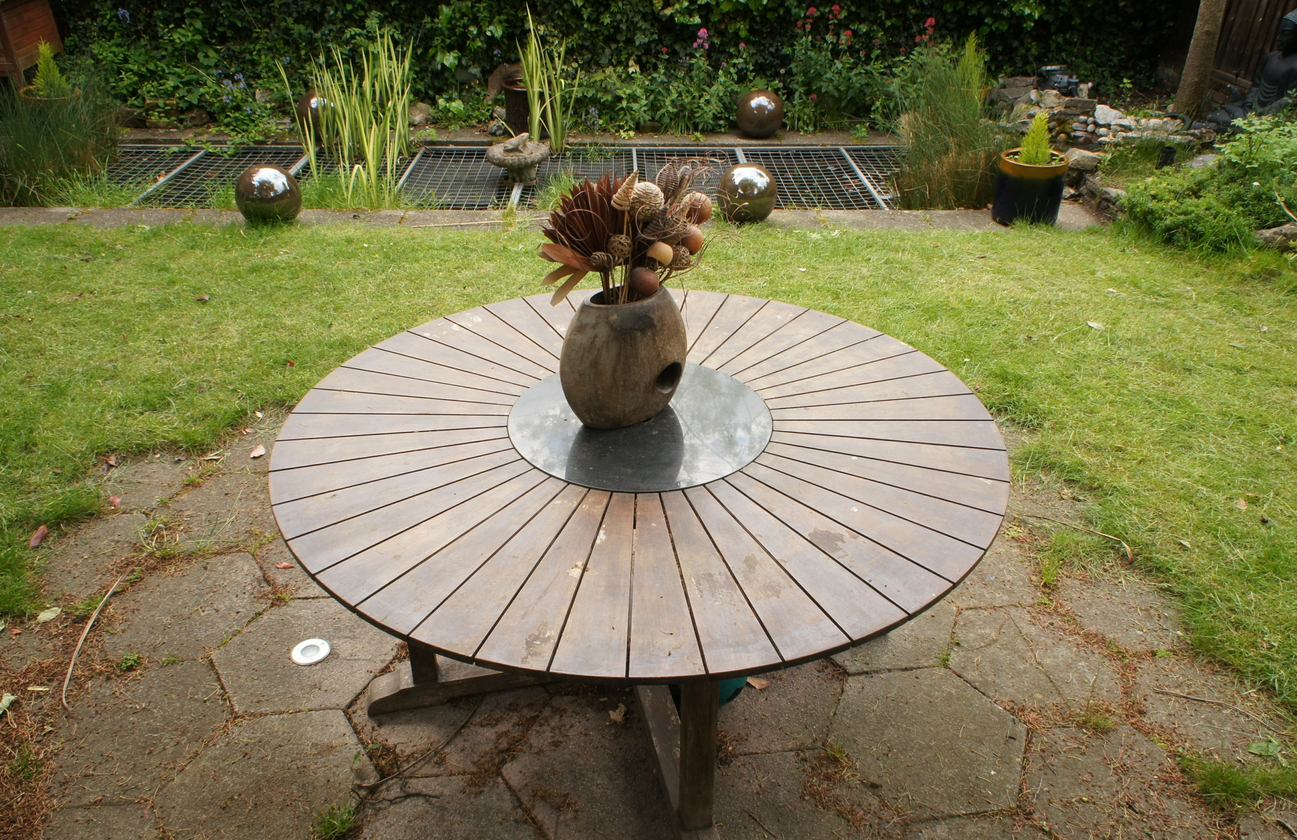}
    \includegraphics[width=0.325\linewidth]{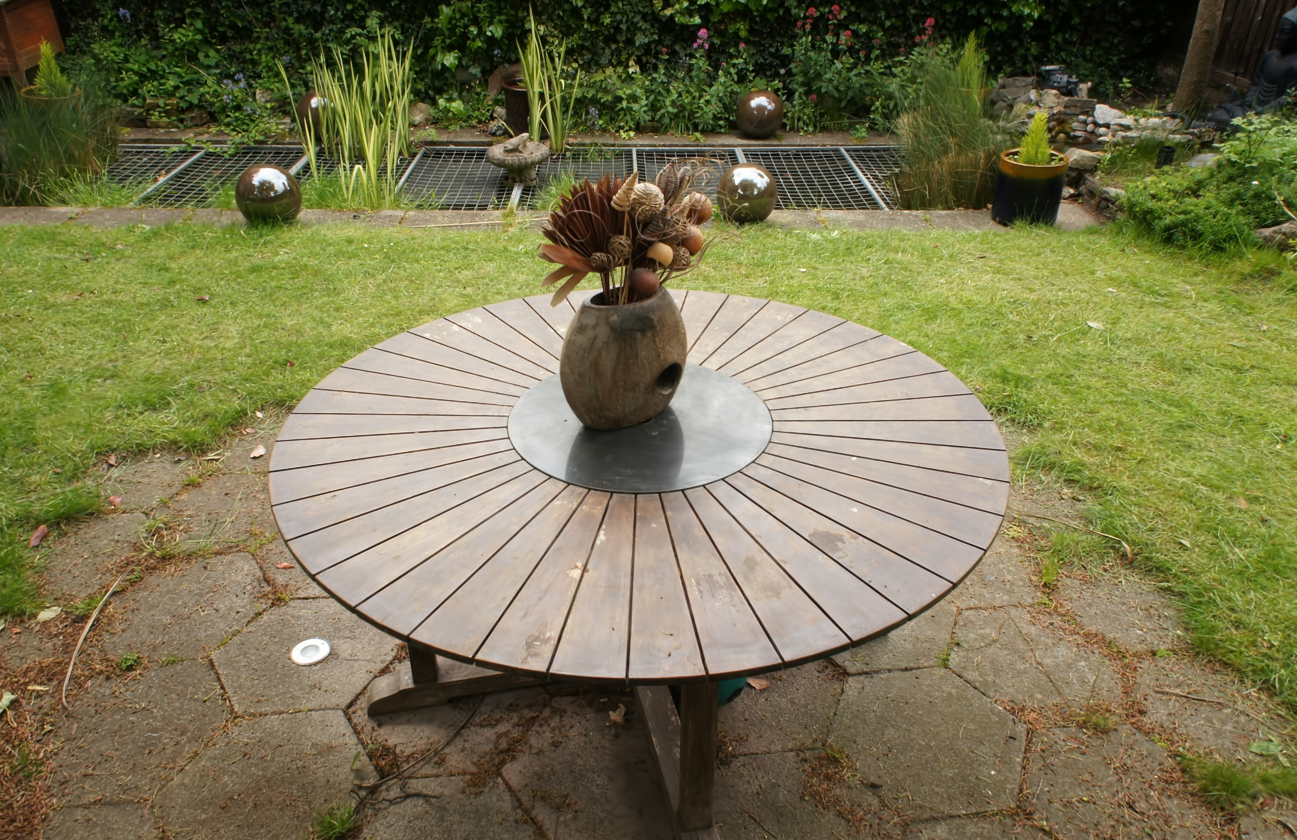}
    \includegraphics[width=0.325\linewidth]{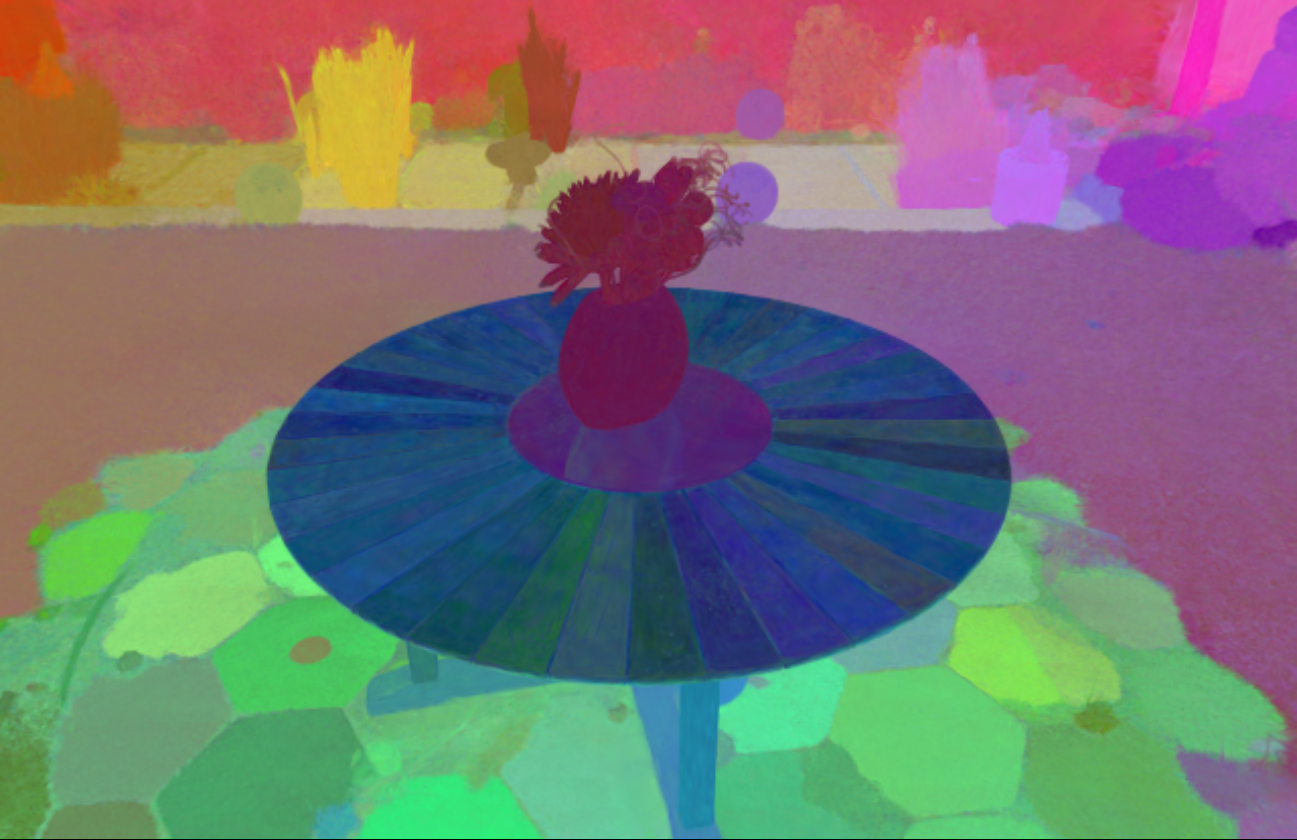}
    \includegraphics[width=0.325\linewidth]{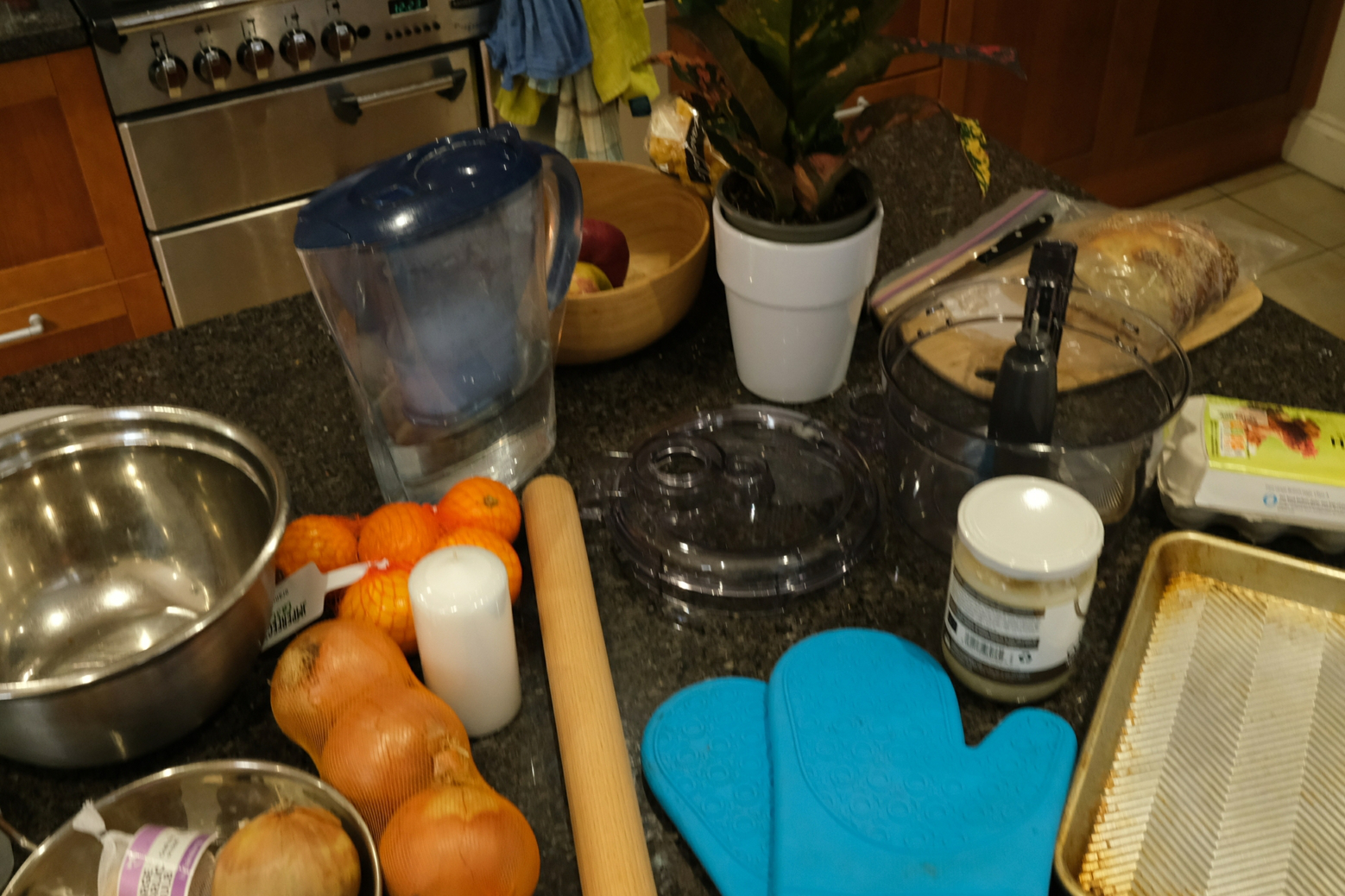}
    \includegraphics[width=0.325\linewidth]{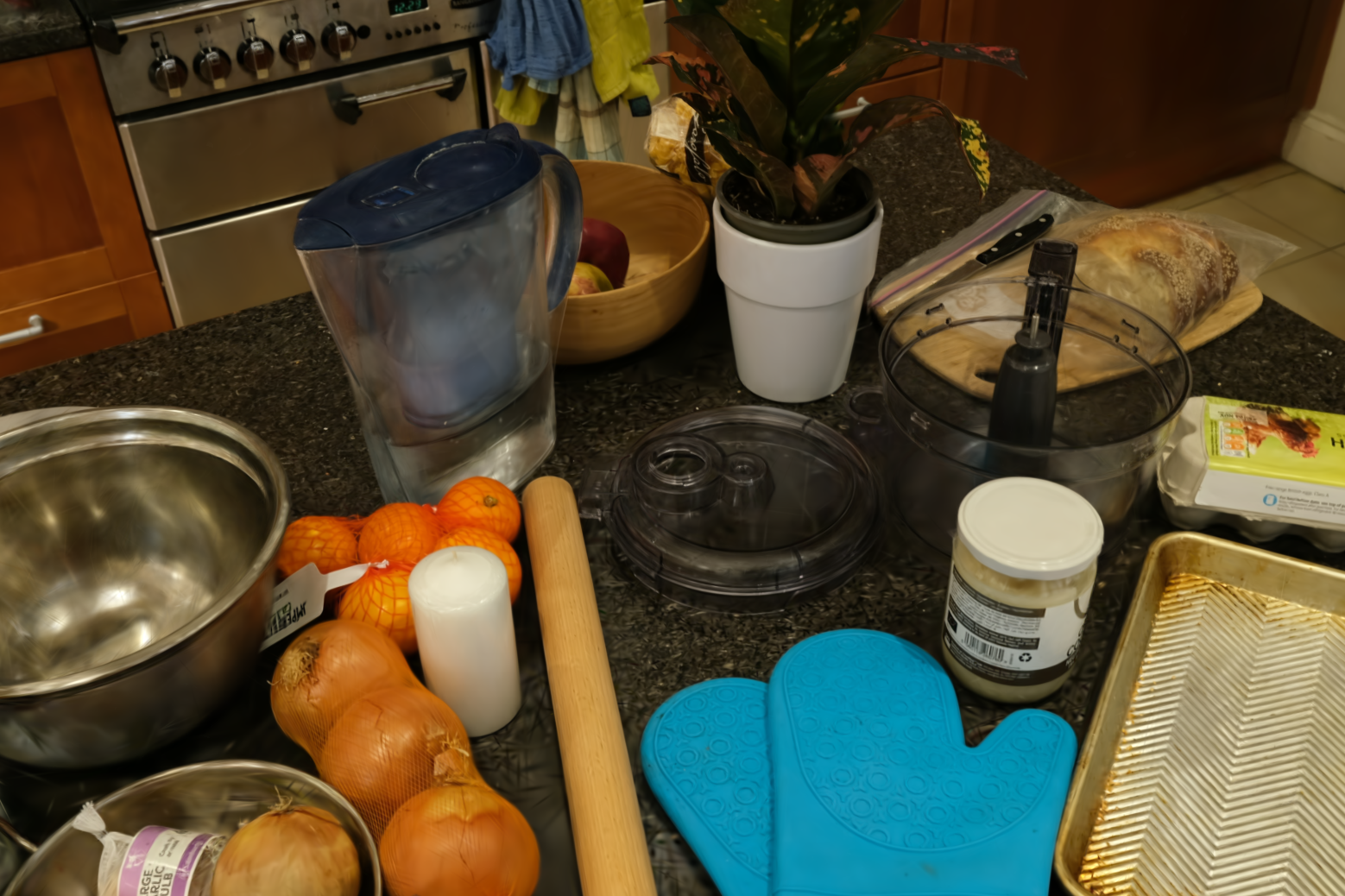}
    \includegraphics[width=0.325\linewidth]{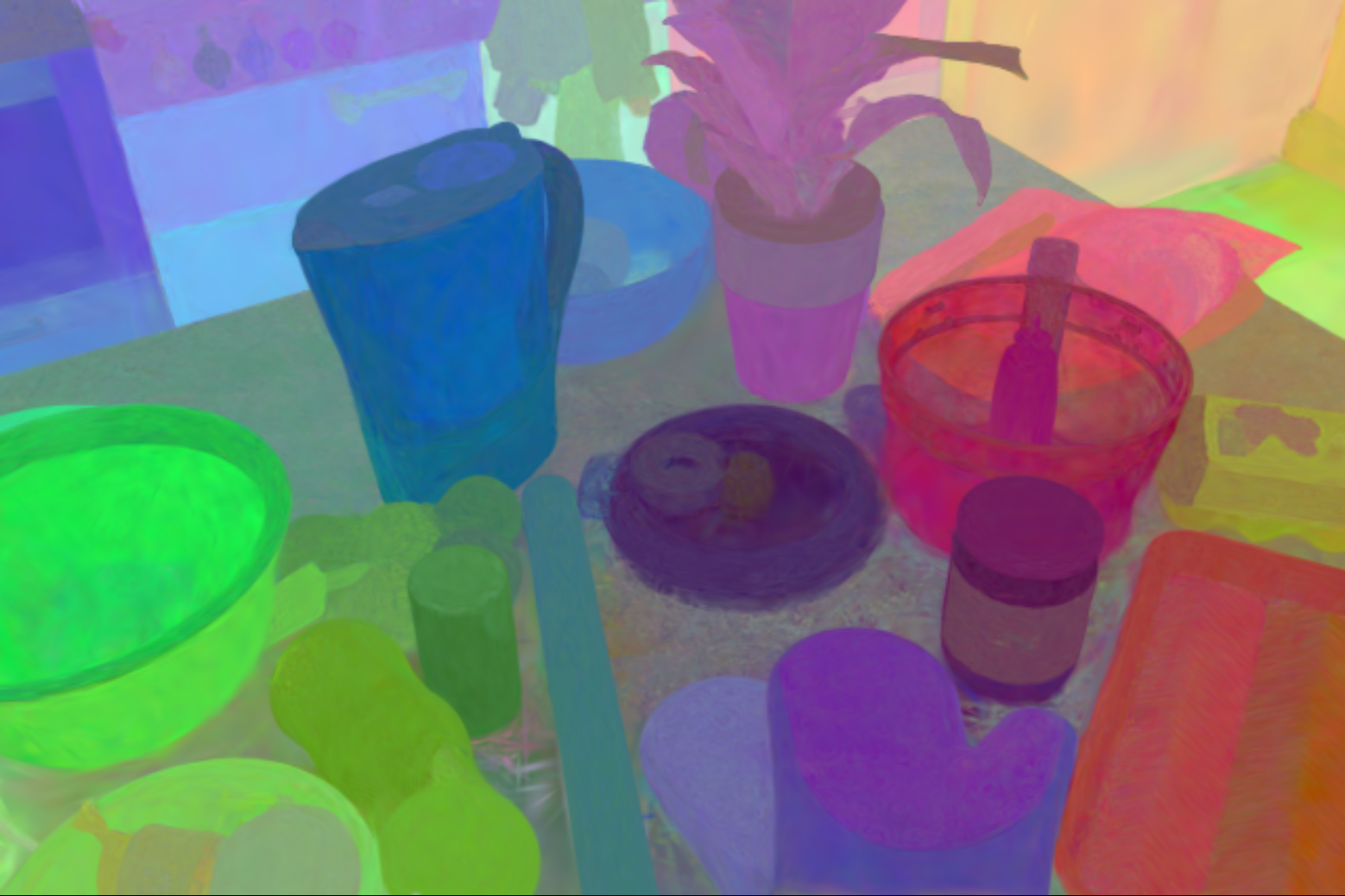}
    \caption{Qualitative results on MipNeRF 360. From left to right: GT image; \coolname~RGB render; \coolname~feature map (PCA).}
    \label{fig:mip360-qual}
\end{figure}

\begin{table}[t]
    \centering
    \begin{adjustbox}{max width=\linewidth}
    \begin{tabular}{c|cccccccc|c}
    \toprule
    Method & Office0 & Office1 & Office2 & Office3 & Office4 & Room0 & Room1 & Room2 & Mean \\
    \midrule
    MVSeg~\cite{mirzaei2023spinnerf} & 31.4 & 40.4 & 30.4 & 30.5 & 25.4 & 31.1 & 40.7 & 29.2 & 32.4 \\
    SA3D~\cite{cen2024sa3dnerf} & 84.4 & 77.0 & 88.9 & 84.4 & 82.6 & 77.6 & 79.8 & 89.2 & 83.0 \\
    OmniSeg3D~\cite{ying2023omniseg3d} & 83.9 & 85.3 & 89.0 & 87.2 & 78.3 & 83.0 & 79.4 & 88.9 & 84.4 \\
    \hline
    \coolname~(Ours) & 82.9 & 78.4 & 85.1 & 84.1 & 80.0 & 77.0 & 85.4 & 84.3 & 82.1 \\
    \bottomrule
    \end{tabular}
    \end{adjustbox}
    \caption{Instance Segmentation results (mean IoU) on Replica dataset. }
    \label{tab:replica-quant}
\end{table}

\begin{figure}[t]
    \centering
    \includegraphics[width=0.325\linewidth]{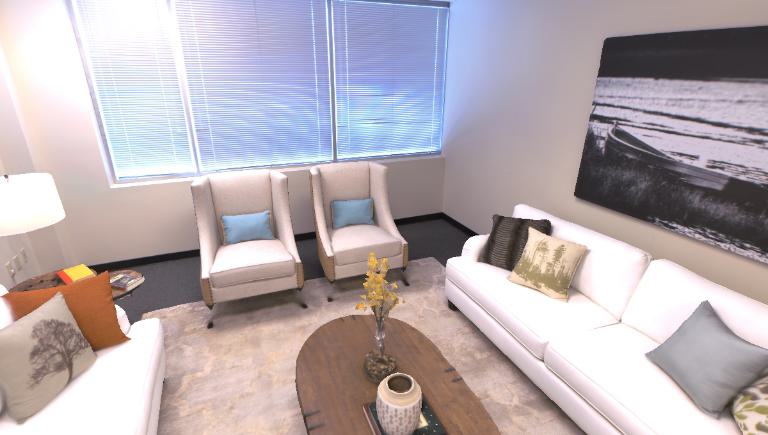}
    \includegraphics[width=0.325\linewidth]{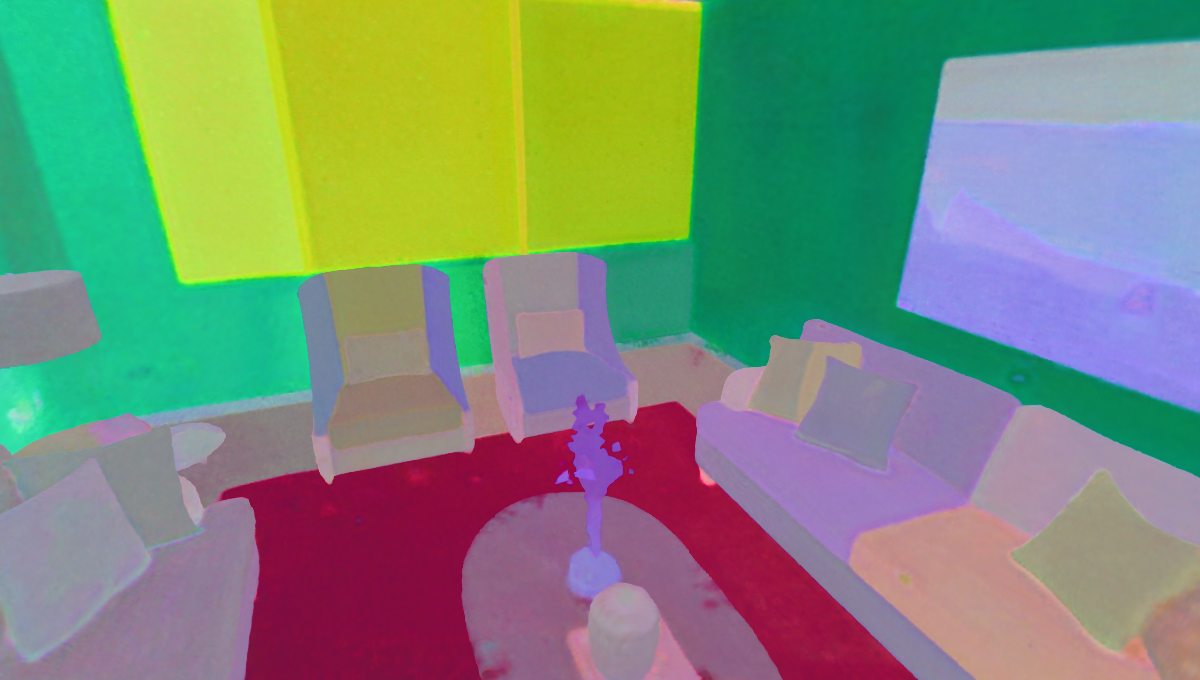}
    \includegraphics[width=0.325\linewidth]{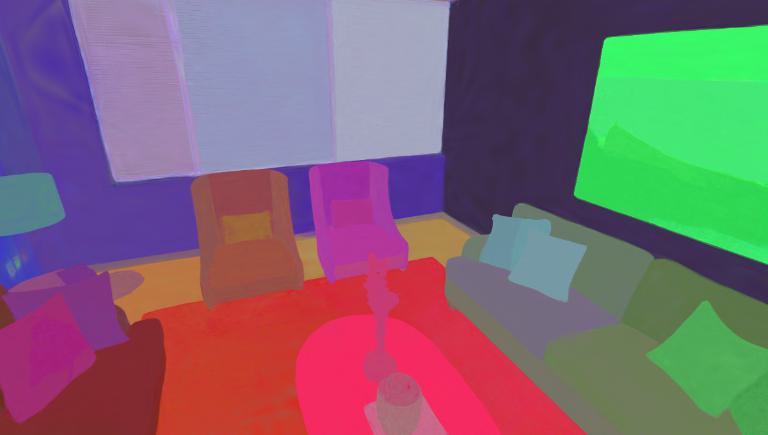}
    \caption{Qualitative result on Replica datasets. From left to right: GT image; OmniSeg3D feature map; \coolname~feature map}
    \label{fig:replica-qual}
\end{figure}

\subsubsection{Subset Splitting}

For sequences that only have a recording from one pair of Aria glasses, the first $4/5$ of the video is considered as seen views and the rest are considered as novel ones. For sequences that have videos from two pairs, the video from one pair is considered as seen views and the other is considered as novel views. 
During training, every 1 out of 5 consecutive frames in the seen views are used for validation the remaining 4 are used for training. The entire novel subset is hidden from training and solely used for evaluation. For evaluation on 2D instance segmentation, we uniformly sampled at most 200 frames from each subset for fast inference. 
The objects in each video sequence are also split into dynamic and static subsets, according to whether their GT object positions have changed by over $2$cm over the duration of each recording. Humans are always considered dynamic objects.

\section{Additional Results}
\label{sec:extra-results}

\subsection{Results on ADT Seen Subset}

For completeness, we also report the 2D instance segmentation and photometric results on the \textbf{seen} subset of ADT in~\cref{tab:adt-2dseg-seen}. Note that the frames used for evaluation in the seen subset are closer to those for training, and therefore these results mostly reflect how well the models overfit the training viewpoints in each scene, rather than generalize to novel views. As we can see from~\cref{tab:adt-2dseg-seen}, \coolname~outperforms the baselines in segmenting static objects using both in-view and cross-view queries. When both static and dynamic objects are considered (the ``All'' column), \coolname~still achieves the best results in cross-view, which is a harder setting for open-world segmentation. \coolname~also has the second place in the in-view setting. 

\subsection{NeRF on ADT Dataset}

In \cref{tab:adt-nerf} and \cref{fig:adt-qual}, we compare \coolname~with the (default) \texttt{nerfacto} model in Nerfstudio~\cite{tancik2023nerfstudio} on the ADT dataset. As we can see from \cref{fig:adt-qual}, although Nerfacto uses per-image appearance embeddings to filter out transient phenomena in reconstruction, it still fails on challenging egocentric datasets like ADT and results in many floaters in the rendering. Quantitatively, \coolname~also outperforms as shown in \cref{fig:adt-qual}. 

\subsection{Non-egocentric public benchmarks}

Non-egocentric benchmarks (Replica, ScanNet, MipNeRF 360) use careful hand-held scanning motions and lack dynamic phenomena. Therefore, they do not reflect the full capability of \coolname. 
We evaluate \coolname~on the MipNeRF 360 dataset~\cite{barron2022mip360} in \cref{tab:mip360-quant} and \cref{fig:mip360-qual}, where we use \coolname-static variant as the scenes are all static. Due to the lack of GT segmentation masks, we provide qualitative results on learned instance features in \cref{fig:mip360-qual}. 
As shown in \cref{tab:mip360-quant} and \cref{fig:mip360-qual}, \coolname~has a similar PSNR as the original 3DGS and learns clean instance features that distinguish different instances. 

We also test~\coolname~on Replica~\cite{replica} and compare to OmniSeg3D~\cite{ying2023omniseg3d}, a recent feature lifting method based on NeRF representation and contrastive learning~\cite{bhalgat2023contrastivelift}. 
We evaluate the instance segmentation task using the multi-view mask propagation protocol~\cite{mirzaei2023spinnerf, cen2024sa3dnerf, ying2023omniseg3d}, where the GT mask from one view is used for computing reference instance features and masks on other view are computed based on the feature distance from the reference ones. 
We follow the evaluation protocol in~\cite{ying2023omniseg3d} and use Eq. (11) in~\cite{ying2023omniseg3d} for computing the similarity scores. 
Similar to the experiments on MipNeRF 360, we used \coolname-static as there is no dynamic content in Replica scenes.

We report the quantitative results (in mIoU) in~\cref{tab:replica-quant} and a qualitative example in~\cref{fig:replica-qual}. 
From~\cref{tab:replica-quant}, we can see that \coolname{} has similar performance with the state-of-the-art NeRF-based segmentation methods~\cite{cen2024sa3dnerf, ying2023omniseg3d} on the non-egocentric Replica dataset. 
From the qualitative example in~\cref{fig:replica-qual}, we can see that \coolname~also results in clean and sharp feature boundaries on Replica as contemporary work OmniSeg3D~\cite{ying2023omniseg3d}, which distinguish different object instances and even the parts within each object.

\section{Additional Discussion on Limitations}

Due to form factor and power constraints, egocentric videos are often captured with more challenges. Due to rapid head motion and lighting condition changes in the egocentric videos, the images contain significant motion blur that causes challenges in recovering sharp reconstructions from them. 
This explains in part the blurry results shown in some of the reconstruction results by \coolname. We leave how to improve the reconstruction quality from egocentric videos for future work.

%
%
\bibliographystyle{splncs04}
\bibliography{main}
\end{document}